\documentclass[lettersize,journal]{IEEEtran}

\usepackage{amssymb,amsmath}
\usepackage{cite}
\usepackage{graphicx}
\usepackage{subfig}

\usepackage{psfrag}
\usepackage{url}
\usepackage[latin1]{inputenc}
\usepackage[absolute,overlay]{textpos}
\usepackage{gensymb}
\usepackage{cases}
\usepackage[font=footnotesize]{caption}
\usepackage{float}
\usepackage[linesnumbered,ruled,lined]{algorithm2e}
\newcommand{\argmax}[1]{{\operatorname{arg}\,\max_{#1}}\,}

\usepackage{pgf}

\usepackage[nocomma]{optidef}

\usepackage{amsthm}

\usepackage{booktabs}
\usepackage{color,soul}

\usepackage{textcomp}
\usepackage{bbm}
\usepackage{xcolor}
\def\BibTeX{{\rm B\kern-.05em{\sc i\kern-.025em b}\kern-.08em
    T\kern-.1667em\lower.7ex\hbox{E}\kern-.125emX}}

\usepackage{tabularx}
\usepackage{makecell}
\usepackage{multirow}

\begin{document}

\title{Conservative and Risk-Aware Offline Multi-Agent Reinforcement Learning}
\author{
	\IEEEauthorblockN{Eslam Eldeeb, Houssem Sifaou, Osvaldo Simeone, Mohammad Shehab, and Hirley Alves
}
	
    \thanks{Eslam Eldeeb and Hirley Alves are with Centre for Wireless Communications (CWC), University of Oulu, Finland. (e-mail: eslam.eldeeb@oulu.fi; hirley.alves@oulu.fi). Houssem Sifaou and Osvaldo Simeone are with the King's Communications, Learning $\text{\&}$ Information Processing (KCLIP) lab within the Centre for Intelligent Information Processing Systems (CIIPS), Department of Engineering, King's College London, WC2R 2LS London, U.K. (e-mail: houssem.sifaou@kcl.ac.uk; osvaldo.simeone@kcl.ac.uk). Mohammad Shehab is with CEMSE Division, King Abdullah University of Science and Technology (KAUST), Thuwal 23955-6900, Saudi Arabia (email: mohammad.shehab@kaust.edu.sa).
    }
    
    \thanks{The work of E. Eldeeb and H. Alves was partially supported by the Research Council of Finland (former Academy of Finland) 6G Flagship Programme (Grant Number: 346208) and by the European Commission through the Hexa-X-II (GA no. 101095759). The work of H. Sifaou and O. Simeone was partially supported by the European Union's Horizon Europe project CENTRIC (101096379). O. Simeone was also supported by the Open Fellowships of the EPSRC (EP/W024101/1) by the EPSRC project (EP/X011852/1), and by Project REASON, a UK Government funded project under the Future Open Networks Research Challenge (FONRC) sponsored by the Department of Science Innovation and Technology (DSIT).}

    \thanks{The second author has contributed to the problem definitions and the experiments. The third author has had an active role in defining the problems, as well as in writing the text, while the last two authors have had a supervisory role and have revised the text.}
}
\maketitle

\begin{abstract}
% Offline reinforcement learning (RL) has emerged as a promising alternative to online RL in applications where interaction with the environment is not an option. This is particularly the case of digital twin (DT)-based wireless networks where a static offline dataset, previously collected from the physical twin, is used to optimize decision-making policies. In this work, we focus on offline multi-agent RL (MARL) and propose a new scheme that is able to handle both the epistemic and aleatoric uncertainties.
%\MS{Title too long}
Reinforcement learning (RL) has been widely adopted for controlling and optimizing complex engineering systems such as next-generation wireless networks. An important challenge in adopting RL is the need for direct access to the physical environment. This limitation is particularly severe in multi-agent systems, for which conventional multi-agent reinforcement learning (MARL) requires a large number of coordinated online interactions with the environment during training. When only offline data is available, a direct application of online MARL schemes would generally fail due to the epistemic uncertainty entailed by the lack of exploration during training. In this work, we propose an offline MARL scheme that integrates distributional RL and conservative Q-learning to address the environment's inherent aleatoric uncertainty and the epistemic uncertainty arising from the use of offline data. We explore both independent and joint learning strategies. The proposed MARL scheme, referred to as multi-agent conservative quantile regression, addresses general risk-sensitive design criteria and is applied to the trajectory planning problem in drone networks, showcasing its advantages.

%Digital twin (DT) platforms are increasingly regarded as a promising technology for controlling, optimizing, and monitoring complex engineering systems such as next-generation wireless networks. An important challenge in adopting DT solutions is their reliance on data collected offline, lacking direct access to the physical environment. This limitation is particularly severe in multi-agent systems, for which conventional multi-agent reinforcement learning (MARL) requires online interactions with the environment. A direct application of online MARL schemes to an offline setting would generally fail due to the epistemic uncertainty entailed by the limited availability of data. In this work, we propose an offline MARL scheme for DT-based wireless networks that integrates distributional RL and conservative Q-learning to address the environment's inherent aleatoric uncertainty and the epistemic uncertainty arising from limited data. To further exploit the offline data, we adapt the proposed scheme to the centralized training decentralized execution framework, allowing joint training of the agents' policies. The proposed MARL scheme, referred to as multi-agent conservative quantile regression (MA-CQR) addresses general risk-sensitive design criteria and is applied to the trajectory planning problem in drone networks, showcasing its advantages.

\end{abstract}

\section{Introduction}\label{sec:introduction}

\subsection{Context and Motivation}
 Recent advances in machine learning (ML) and artificial intelligence (AI), high-performance computing, cloudification, and simulation intelligence~\cite{lavin2021simulation} have supported the development of data-driven paradigms for the engineering of complex systems, such as wireless networks~\cite{wang2020artificial,chen2019artificial}. Reinforcement learning (RL) is a particularly appealing methodology for settings requiring dynamic decision-making, for which feedback can be distributed automatically or via human judgement~\cite{8714026,chen2021deep}. An important challenge in adopting RL solutions is their reliance on online interaction with the environment. This limitation is particularly severe in multi-agent systems, for which conventional multi-agent reinforcement (MARL) requires a large number of coordinated online interactions with the environment~\cite{marl-book}.

%: \emph{digital twin} (DT) platforms  ~\cite{saracco2019digital,raza2020digital}. A DT system consists of a digital mirror of the physical twin (PT) counterpart built and maintained using data from the PT. Originating in manufacturing, DTs have emerged as a promising solution for fields such as healthcare and next-generation wireless networks, allowing algorithms and ML models to be optimized and tested virtually before deployment on the PT~\cite{coveney2023virtual,khan2022digital}. In the context of wireless networks, DTs are particularly well suited as a component in open radio access network architectures (RANs)~\cite{mirzaei2023network,villa2023colosseum}. %\MS{In myh opinion mentioning ORANs here as a buzz without further details is quite clumsy. Either remove it or mention more details}.

\begin{figure}[t!]
    \centering    \includegraphics[width=1\columnwidth,trim={1.5cm 0 1.5cm 0},clip]{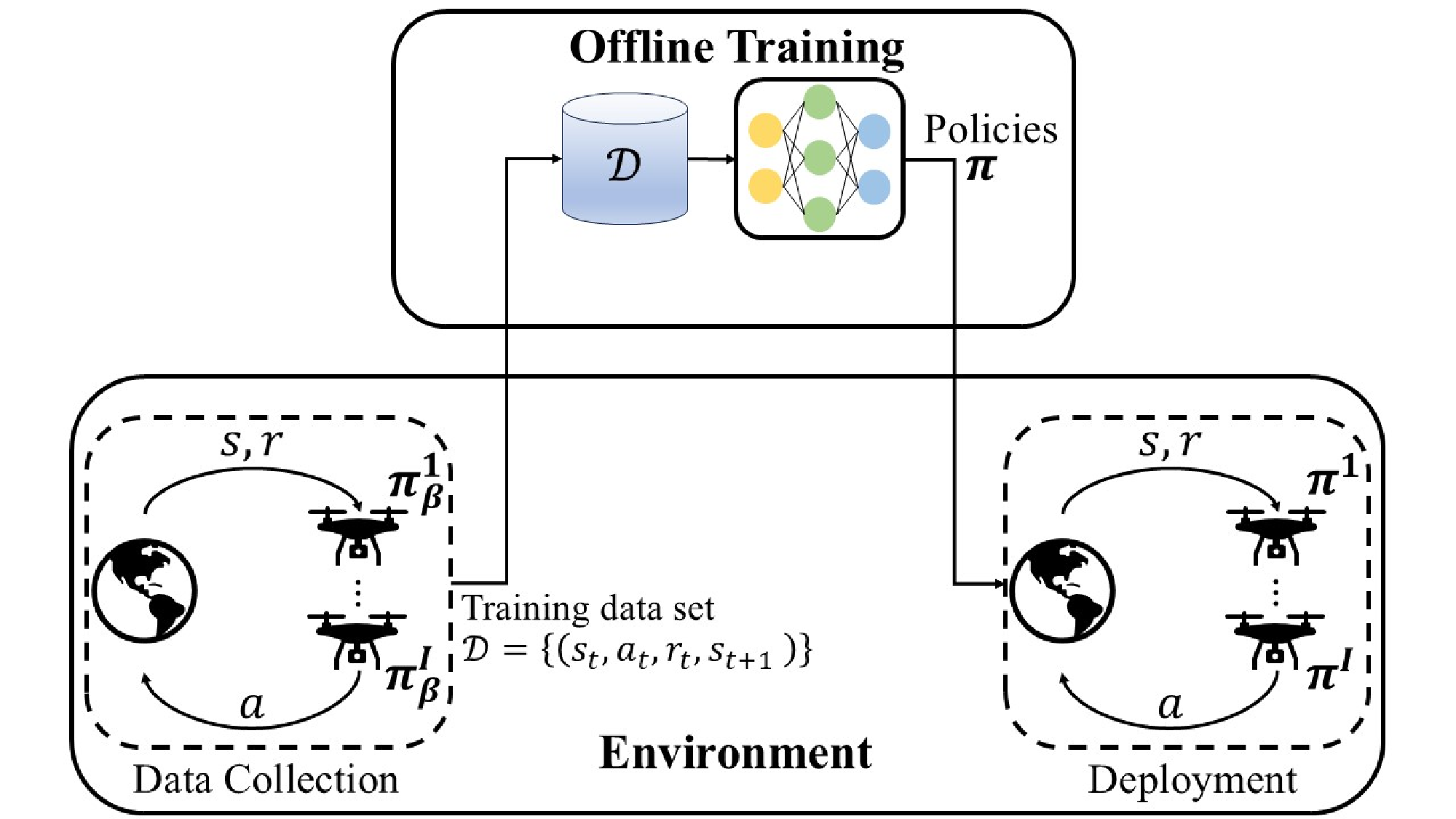} \vspace{2mm}
    \caption{Consider access to data collected offline following some fixed and unknown policies $\pi_{\beta} = \{\pi_{\beta}^i\}^I_{i=1}$ in an environment consisting of $I$ agents. Based on this dataset, the goal is to optimize policies $\pi = \{\pi^i\}^I_{i=1}$ for the agents while ensuring robustness to the  uncertainty arising from the stochastic environment, from the limited data, and from the lack of interactions with the environment.}
    \vspace{0mm}
    \label{Online_Offline_Fig}
\end{figure}

%The DT generally updates its internal representation of the PT and the ML models being trained on behalf of the PT based on data received from the PT. The process of synchronizing the DT is limited by the communication bottleneck between DT and PT~\cite{villa2023colosseum,zheng2023data}. As a result, an important challenge in adopting DT solutions is their reliance on data collected offline, lacking direct access to the physical system. This limitation is particularly severe in multi-agent systems, for which conventional multi-agent reinforcement (MARL) requires online interactions with the environment \cite{marl-book}.%`
% {\color{blue} add a basic introduction to MARL, preferably a book}. 

% \textcolor{blue}{"The key enablers of future wireless networks, such as digital twins, aerial networks, radio resource management, and intelligent meta-surface, is motivated multi-agent systems~\cite{feriani2021single}. In addition, utilizing multiple agents in wireless systems improves the learning performance of these systems\cite{li2022applications,9815722}. }

When only data collected offline is available, a direct application of online MARL schemes would generally fail due to the \emph{epistemic uncertainty} entailed by the limited availability of data. In particular, even in the case of a single agent, \emph{offline reinforcement learning}, which relies only on offline data, may over-estimate the quality of given actions that happened to perform well during data collection due to the inherent stochasticity and outliers of the environment \cite{levine2020offline}.
% {\color{blue} ref to paper with simple example}\cite{levine2020offline}. 
This problem can be addressed in online RL via exploration, trying actions, and modifying return estimates based on environmental feedback. However, as mentioned, exploration is not feasible in offline RL, as policy design is based solely on the offline dataset. Furthermore, in multi-agent systems, this problem is exacerbated by the inherent uncertainty caused by the non-stationary behavior of other agents during training \cite[Chapter 11]{bdr2023}.

In this paper, we propose a novel offline MARL strategy, multi-agent conservative quantile regression (MA-CQR), that addresses the overall uncertainty caused by the use of offline data. The introduced approaches are termed \emph{multi-agent
conservative independent quantile regression} (MA-CIQR) via independent learning and \emph{multi-agent
conservative centralized quantile regression} (MA-CCQR) via joint training. These {approaches} integrate distributional RL \cite{bdr2023} and offline RL \cite{kumar2020conservative} to support a risk-sensitive multi-agent design that mitigates impairments arising from access to limited data from the environment. We showcase the performance of MA-CIQR and MA-CCQR by focusing on the problem of designing trajectories of unmanned aerial vehicles (UAVs) used to collect data from sensors in an Internet-of-Things (IoT) scenario \cite{abd2019deep,samir2020age,eldeeb2022multi} (see Fig.~\ref{Online_Offline_Fig}). It is noted that this work considers the worst case in which agents can only use offline data without relying on an internal model of the environment. Future work may investigate settings in which the agents have prior information about the environment that can be used to learn a world model via methods such as model-based offline RL~\cite{kidambi2020morel} (see Sec. I.B for a review).

\subsection{Related Work}
\label{Lit_revi}

\noindent\emph{Offline RL:} Offline RL has gained increasing interest in recent years due to its wide applicability to domains where online interaction with the environment is impossible or presents high costs and risks. Offline RL relies on a static offline transition dataset collected from the environment using some \emph{behavioral} policy. The behavioral policy is generally suboptimal and may be unknown to the designer~\cite{levine2020offline}. The reliance on a suboptimal policy for data collection distinguishes offline RL from imitation learning, in which the goal is reproducing the behavior of an expert policy \cite{ciosek2022imitation}. The discrepancy between the behavior and optimized policies creates a \emph{distributional shift} between training data and design objective. This shift could be resolved by collecting more data, but this is not possible in an offline setting. Therefore, the distributional shift contributes to the epistemic uncertainty of the agent.

Several approaches have been proposed to address this problem in offline RL. One class of methods constrains the difference between the learned and behavior policies~\cite{schulman2015trust}. Another popular approach is to learn conservative estimates of the action-value function or Q-function. Specifically, \emph{conservative Q-learning} (CQL), proposed in~\cite{kumar2020conservative}, penalizes the values of the Q-function for \emph{out-of-distribution} (OOD) actions. OOD actions are those whose impact is not sufficiently covered by the dataset. 

Other works have leveraged offline RL with model-based RL by exploiting information about the physical environment~\cite{kidambi2020morel,yu2020mopo,Barde2024}. The work in~\cite{kidambi2020morel} proposed a model-based offline reinforcement learning framework that first learns the transition dynamics of the environment from the offline data and then optimizes the policy. To address the distributional shift arising in offline RL, reference~\cite{yu2020mopo} modified conventional model-based RL schemes by penalizing the rewards by the amount of uncertainty regarding the environment dynamics.
In~\cite{Barde2024}, a model-based solution for offline MARL was developed for coordination-insensitive settings. The approach consists of learning a world model from the dataset and using it to optimize the agents' policies.

Regarding applications of offline RL to wireless systems, the recent work~\cite{yang2023offline} investigated a radio resource management problem by comparing the performance of several \emph{single-agent} offline RL algorithms.

\noindent \emph{Distributional RL}:
Apart from offline RL via CQL, the proposed scheme builds on \emph{distributional RL} (DRL), which is motivated by the inherent aleatoric uncertainty caused by the stochasticity of the environment~\cite{bellemare2017distributional,dabney2017distributional,dabney2018implicit}. Rather than targeting the average return as in conventional RL, DRL maintains an estimate of the \emph{distribution} of the return. This supports the design of \emph{risk-sensitive} policies that disregard gains attained via risky behavior, favoring policies that ensure satisfactory worst-case performance levels instead. 

A popular risk measure for use in DRL is the \emph{conditional value at risk} (CVaR)~\cite{rockafellar2000optimization,lim2022distributional,dabney2018implicit,ma2021conservative}, which evaluates the average performance by focusing only on the lower tail of the return distribution. Furthermore, a state-of-the-art DRL strategy is \emph{quantile-regression deep Q-network} (QR-DQN), which approximates the return distribution by estimating $N$ uniformly spaced quantiles~\cite{dabney2017distributional}.

% We investigate in this work the problem of distributional shift for offline MARL in the context of DT-based wireless systems.#
%\noindent \emph{Offline MARL}: 
%{\color{blue} please make sure to distinguish independent per-agent strategies and centralized training decentralized execution strategies such as https://arxiv.org/abs/2307.11620. Also make sure to cite references [11] and [12] in that paper.} 
%Ref 11 in that paper~\cite{xu2023offline} [RL] and Ref 12 [MARL] in that paper~\cite{yang2021believe}.
\noindent \emph{Offline MARL}: Recently, several works have been proposed that adapt the idea of conservative offline learning to the context of multi-agent systems (offline  MARL)~\cite{jiang2023offline,pan2022plan,shao2023counterfactual,wang2023offline,yang2021believe}. Specifically, conservative estimates of the value function in a decentralized fashion are obtained in~\cite{jiang2023offline} via value deviation and transition normalization. Several other works proposed centralized learning approaches. The authors in~\cite{pan2022plan} leveraged first-order policy gradients to calculate conservative estimates of the agents' value functions. The work~\cite{shao2023counterfactual} presented a counterfactual conservative approach for offline MARL, while~\cite{wang2023offline} introduced a framework that converts global-level value regularization into equivalent implicit local value regularization. The authors in~\cite{yang2021believe} addressed the overestimation problem using implicit constraints.

Overall, all of these works focused on \emph{risk-neutral} objectives, hence not making any provisions to address risk-sensitive criteria.  In this regard, the paper~\cite{ma2021conservative} combined distributional RL and conservative Q-learning to develop a risk-sensitive algorithm, but only for single-agent settings.

\noindent\emph{Applications of MARL to wireless systems:} Due to the multi-objective and multi-agent nature of many control and optimization problems in wireless networks, MARL has been adopted as a promising solution in recent years. For instance, related to our contribution, the work in~\cite{eldeeb2023traffic} proposed an online MARL algorithm to jointly minimize the age-of-information (AoI) and the transmission power in IoT networks with traffic arrival prediction, whereas the authors in~\cite{9322539} leveraged MARL for AoI minimization in UAV-to-device communications. Moreover, MARL was used in~\cite{8807386} for resource allocation in UAV networks. The work~\cite{8792117} developed a MARL-based solution for optimizing power allocation dynamically in wireless systems. The authors in~\cite{naderializadeh2021resource} used MARL for distributed resource management and interference mitigation in wireless networks and in~\cite{xu2023digital}, edge-end task division, transmit power, computing resource type matching and allocation are jointly optimized using a MARL algorithm.

\noindent\emph{Applications of Distributional RL to wireless systems:} Distributional RL has been recently leveraged in \cite{zhang2021millimeter} to carry out the optimization for a downlink multi-user communication system with a base station assisted by a reconfigurable intelligent reflector (IR). Meanwhile, reference~\cite{zhang2020distributional} focused on the case of mmWave communications with IRs on a UAV. Distributional RL has also been used in~\cite{hua2019gan} for resource management in network slicing. The paper~\cite{ma2021conservative} combined distributional RL and conservative Q-learning to develop a risk-sensitive algorithm, but only for single-agent settings.

All in all, to the best of our knowledge, our work in this paper is the first to integrate conservative offline RL and distributional MARL, and it is also the first to investigate the application of offline MARL to wireless systems.

%\MS{The next sentence is out of context. I suggest removing it. The introduction is too long already.}

%\noindent \emph{Applications of distributional RL to wireless systems}: 
% Lastly, one of the latest works on offline RL in wireless is introduced in~\cite{yang2023offline}, where the authors present an offline RL scheme for radio resource management (RRM).

 \subsection{Main Contributions}

 This work introduces MA-CQR, a novel offline MARL scheme that supports optimizing risk-sensitive design criteria such as CVaR. MA-CQR is evaluated on the relevant problem of UAV trajectory design for IoT networks. The contributions of this paper are summarized as follows. 
\begin{itemize}
    \item We propose MA-CQR, a novel conservative and distributional offline MARL solution. MA-CQR leverages quantile regression (QR) to support the optimization of risk-sensitive design criteria and CQL to ensure robustness to OOD actions. As a result, MA-CQR addresses both the epistemic uncertainty arising from the presence of limited data and the aleatoric uncertainty caused by the randomness of the environment. 
    \item We present two versions of MA-CQR with different levels of coordination among the agents. In the first version, referred to as MA-CIQR, the agents' policies are optimized independently. In the second version, referred to as MA-CCQR, we leverage value decomposition techniques that allow joint training~\cite{sunehag2017value,lyu2024centralizedcriticsmultiagentreinforcement}.
    
    \item To showcase the proposed schemes, we consider a trajectory optimization problem in UAV networks~\cite{eldeeb2023traffic}. As illustrated in Fig. 1, the system comprises multiple UAVs collecting information from IoT devices. The multi-objective design tackles the minimization of the AoI for data collected from the devices and the overall transmit power consumption. We specifically exploit MA-CQR to design risk-sensitive policies that avoid excessively risky trajectories in the pursuit of larger average returns. 
    \item Numerical results demonstrate that MA-CIQR and MA-CCQR versions yield faster convergence and higher returns than the baseline algorithms. Furthermore, both schemes can avoid risky trajectories and provide the best worst-case performance. Experiments also depict that centralized training provides faster convergence and requires less offline data.
\end{itemize}

The rest of the paper is organized as follows. Section~\ref{sec:problem_def} describes the  MARL setting and the design objective. Section~\ref{sec:background} introduces distributional RL and conservative Q-Learning. In section~\ref{sec:proposed}, we present the proposed MA-CIQR algorithm using independent training, whereas section~\ref{sec:proposed_Cent} presents the proposed MA-CCQR algorithm using centralized training. In Section~\ref{sec:results}, we provide numerical experiments on trajectory optimization in UAV networks. Section~\ref{sec:conclusions} concludes the paper.

%\MS{I suggest adding a table here to summarize abbreviations and symbols}

\begin{table}[t!]
\centering
\caption{Abbreviations}
\label{Abbrev}
\begin{tabular}{|l|l|}
\hline
AoI & Age-of-information \\ \hline
CDF & Cumulative distribution function \\ \hline
CQL & Conservative Q-learning \\ \hline
%CTDE & Centralized training and decentralized execution \\ \hline
CVaR & Conditional value at risk \\ \hline
DQN & Deep Q-network \\ \hline
DRL & Distributional reinforcement learning \\ \hline
%DT & Digital twin \\ \hline
MA-CCQL & Multi-agent conservative centralized Q-learning \\ \hline
MA-CCQR & Multi-agent conservative centralized quantile regression \\ \hline
MA-CIQL & Multi-agent conservative independent Q-learning \\ \hline
MA-CIQR & Multi-agent conservative independent quantile regression \\ \hline
MA-CQL & Multi-agent conservative Q-learning \\ \hline
MA-CQR & Multi-agent conservative quantile regression \\ \hline
MA-DCQN & Multi-agent deep centralized Q-network \\ \hline
MA-DIQN & Multi-agent deep independent Q-network \\ \hline
MA-DQN & Multi-agent deep Q-network \\ \hline
MA-QR-DCQN & Multi-agent quantile regression deep centralized Q-network \\ \hline
MA-QR-DIQN & Multi-agent quantile regression deep independent Q-network \\ \hline
MA-QR-DQN & Multi-agent quantile regression deep Q-network \\ \hline
MARL & Multi-agent reinforcement learning \\ \hline
OOD & Out-of-distribution \\ \hline
%PT & Physical twin \\ \hline
QR-DQN & Quantile-regression deep Q-network \\ \hline
UAV & Unmanned aerial vehicles\\ 

\hline
\end{tabular} 
\end{table}

\begin{table}[t!]
\centering
\caption{Notations}
\label{Notations}
\begin{tabular}{|l|l|}
\hline
$I$ & Number of agents \\ \hline

$s_t$ & Overall state of the environment at time step $t$ \\ \hline

$a_t$ & Joint action of all agents at time step $t$ \\ \hline

$a_t^i$ & Action of agent $i$ at time step $t$ \\ \hline

$r_t$ & Immediate reward at time step $t$\\ \hline

$\gamma$ & Discount factor \\ \hline

$Q(s,a)$ & Q-function \\ \hline

$Z(s,a)$ & Return starting from $(s,a)$ \\ \hline

$P(s_{t+1} | s_t,a_t)$ & Transition probability \\ \hline

$\pi^i(a^i_t \mid s_t )$ & Policy of agent $i$ \\ \hline

$P_{Z(s,a)}$ & Distribution of the return \\ \hline

%$P({\mathcal{T}})$ & Trajectory distribution \\ \hline

$R (r_t \mid s_t,a_t)$ & Stationary reward distribution \\ \hline

$\xi$ & Risk tolerance level \\ \hline

$J_{\xi}^{\text{CVaR}}$ & CVaR risk measure \\ \hline

$F_{Z^\pi}^{-1}\big(\xi\big)$ & Inverse CDF of the return \\ \hline

$\mathcal{D}$ & Offline dataset collected \\ \hline

%$\mathcal{L} (Q^i, \hat{Q}^{i(k)})$ & Offline DQN loss of agent $i$\\ \hline

$\theta_j^i(s,a)$ & Quantile estimate of the distribution $
P_{Z^i(s,a)}(\theta^i)$ \\ \hline

$\zeta_{\tau}^{}(u)$ & Quantile regression Huber loss \\ \hline

$\Delta^{i(k)}_{j j^{\prime}}$ & TD errors evaluated with the quantile estimates \\ &of agent $i$ \\ \hline

$\alpha$ & CQL hyperparameter \\ \hline

$M$ & Number of devices in the system \\ \hline

$A_t^m$ & AoI of device $m$ at time step $t$ \\ \hline

$g_t^{i,m}$ & Channel gain between agent $i$ and device $m$ \\ &at time step $t$ \\ \hline

$P_t^{i,m}$ & Transmission power for device $m$ to communicate with \\ &agent $i$ at time step $t$ \\ \hline

$p_\mathrm{risk}$ & Risk probability \\ \hline

$\mathrm{P_\mathrm{risk}}$ & Risk penalty \\

\hline
\end{tabular} 
\end{table}

\section{Problem Definition}\label{sec:problem_def}
In this section, we describe the multi-agent setting and formulate the problem. This discussion will be also instrumental in introducing the necessary notation, which will be leveraged in the following section to introduce important background information. We consider the setting illustrated in Fig.~\ref{Online_Offline_Fig}, where $I$ agents act in a \emph{physical environment} that evolves in discrete time as a function of the agents' actions and random dynamics. The design of the agents' policies $\pi = \{\pi^i\}_{i=1}^I$ is carried out at a central unit in Fig. \ref{Online_Offline_Fig} that has only access to a fixed dataset $\mathcal{D}$, while not being able to interact with the physical system. The dataset $\mathcal{D}$ is collected offline by allowing the agents to act in the environment according to arbitrary, fixed, and generally unknown policies $\pi_\beta = \{\pi_\beta^i\}_{i=1}^I$. In this section, we describe the multi-agent setting and formulate the offline learning problem. Tables~\ref{Abbrev} and \ref{Notations} summarize the list of abbreviations and notations.

\subsection{Multi-Agent Setting}
Consider an environment characterized by a time-variant state $s_t$, where $t=1,2,...$ is the discrete time index. At time step $t$, each agent $i$ takes \emph{action} $a^i_t \in \mathcal{A}^i$ within some discrete action space $\mathcal{A}^i$. We denote by $a_t = \big[ a_t^1, \cdots, a_t^I \big]$ the vector of actions of all agents at timestep $t$. The state $s_t$ evolves according to a \emph{transition probability} $P(s_{t+1} | s_t,a_t)$ as a function of the current state $s_t$ and of the action vector $a_t$. The transition probability $P(s_{t+1} | s_t,a_t)$ is stationary, i.e., it does not vary with time index $t$.

We focus on a \emph{fully observable multi-agent reinforcement learning} setting, in which each agent $i$ has access to the full system state $s_t$ and produces action $a_t^i$ by following a \emph{policy} $\pi^i(a^i_t \mid s_t )$. 
% All agents receive a common reward $r_t$ distributed as $R\big(r_t \mid s_t, a_t\big)$.
\subsection{Design Goal}

The desired goal is to find the optimal policies $\pi_*(a | s) = \{ \pi_*^i( a | s ) \}_{i=1}^{ I}$ that maximize a \textit{risk measure} $\rho(\cdot)$ of the \textit{return} $Z^{\pi} = \sum_{t=0}^\infty \gamma^t r_t$, which we write as
\begin{equation}
\label{MARL_Objective}
    J_{\rho}(\pi) = \rho_{}  \left[ Z^{\pi}\right],
\end{equation}
where $0 < \gamma < 1$ is  a given discount factor. The distribution of the return $Z^{\pi}$ depends on the policies $\pi$ through the distribution of the trajectory $\mathcal{T} = (s_0, a_0, r_0, s_1,a_1,r_1, ...)$, which is given by
\begin{equation}
\label{traj_dist}
P({\mathcal{T}}) = P(s_0)
\prod_{t=0}^\infty \pi(a_t \mid s_t) R (r_t \mid s_t,a_t) P(s_{t+1} \mid s_t,a_t),
\end{equation}
with $\pi(a_t \mid s_t)=\prod_{i=1}^I \pi^i(a_t^i \mid s_t)$ being the joint conditional distribution of the agents' actions; $P(s_0)$ being a fixed initial distribution; and  $R (r_t \mid s_t,a_t)$ being the stationary \emph{reward distribution}.

The standard choice for the risk measure in~\eqref{MARL_Objective} is the expectation $\rho[\cdot] = \mathbb{E}[\cdot]$, yielding the standard criterion
\begin{equation}
\label{MARL_Objective_Average}
    J^{\text{avg}}(\pi) = \mathbb{E} \big[Z^{\pi} \big].
\end{equation}
The average criterion in~\eqref{MARL_Objective_Average} is considered to be \emph{risk neutral}, as it does not directly penalize worst-case situations, catering only to the average performance.

In stochastic environments where the level of aleatoric uncertainty caused by the transition probability and/or the reward distribution is high, maximizing the expected return may not be desirable since the return $Z^{\pi}$ has high variance. In such scenarios, designing \emph{risk-sensitive} policies may be preferable to enhance the worst-case outcomes while reducing the average performance~\eqref{MARL_Objective_Average}.

\begin{figure}[t!]
    \centering    \includegraphics[width=1\columnwidth,trim={0 0 0 0},clip]{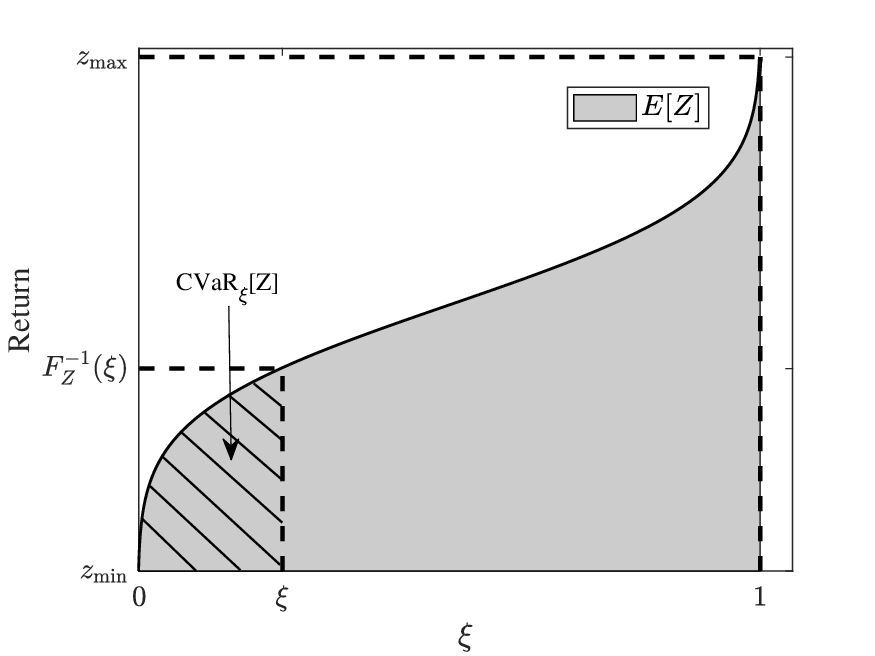} \vspace{2mm}
    \caption{Illustration of the conditional value-at-risk (CVaR). The quantile function $F_Z^{-1}(\xi)$ is plotted as a function of the risk tolerance level $\xi$. The shaded area representing the lower tail of the distribution depicts the $\xi$-level CVaR.} 
    \vspace{0mm}
    \label{CVaR_Ilustration_Fig}
\end{figure}

A common risk-sensitive measure is the \emph{conditional value-at-risk} (CVaR)~\cite{rockafellar2000optimization}, which is defined as the conditional mean 
\begin{equation}
\label{CVaR_Risk}
    J_{\xi}^{\text{CVaR}}(\pi) = \mathbb{E} \big[Z^{\pi} \mid Z^{\pi} \leq F_{Z^\pi}^{-1}\big(\xi\big) \big],
\end{equation}
where $F_{Z^\pi}^{-1}\big(\xi\big)$ is the inverse cumulative distribution function (CDF) of the return ${Z^\pi}$ for some $\xi \in [0,1]$, i.e., the $\xi$-th quantile of the distribution of the return. The CVaR, illustrated in Fig. \ref{CVaR_Ilustration_Fig}, focuses on the lower tail of the return distribution by neglecting values of the return that are larger than the $\xi$-th quantile $F_{Z^\pi}^{-1}\big(\xi\big)$. Accordingly, the probability $\xi$ represents the \textit{risk tolerance level}, with $\xi = 1$ recovering the risk-neutral objective \eqref{MARL_Objective_Average}. The CVaR can also be written as the integral of the quantile function $ F_{Z^{\pi}}^{-1}(\xi)$ as
\begin{equation}
\label{CVaR_obj}
J_\xi^{\text{CVaR}}(\pi) = \frac{1}{\xi} \int_0^{\xi} F_{Z^{\pi}}^{-1}(u) \text{d}u.
\end{equation}
% be te quantile function, which returns $\tau^{th}$ quantile of the distribution of the random variable $Z$. Thus, we have the inequality $P\big(z \leq F_Z^{-1}(\tau) \big) \geq \tau$. The function $F_Z^{-1}\big(\tau\big)$ is monotonically increasing, from the minimum return for $\tau = 0$, to the maximum return for $\tau = 1$, and indicates the maximum return level that is allowed in the fraction $\tau$ of the worst trajectories of the system. The conditional value-at-risk (CVaR) is a risk measure, at which the conditional average
% \begin{equation}
% \label{CVaR_Risk}
%     J_{\tau}^{CVaR}(\pi) = \mathbb{E} \big[z \mid z \leq F_Z^{-1} (\tau) \big],
% \end{equation}
% where we condition on. With $\tau = 1$, the criterion becomes risk-neutral and it reduces to the average measure in~\eqref{MARL_Objective_Average}. Mathematically, the CVaR objective $J^{CVaR}(\pi)$ can be expressed as the integral
% \begin{equation}
% \label{CVaR_obj}
% J^{CVaR}(\pi) = \frac{1}{\tau} \int_0^{\tau} F_Z^{-1}(u) du.
% \end{equation}
% To facilitate the design of risk-sensitive policies, such as CVaR-based policies, the return distribution can be learned using the recently developed distributional RL framework~\cite{bdr2023,bellemare2017distributional,dabney2017distributional}, which will be introduced in the next section. 

\subsection{Offline Multi-Agent Reinforcement Learning}
Conventional MARL~\cite{lowe2020multiagent} assumes that agents optimize their policies $\pi = \{ \pi^i ( a \mid s ) \}_{i =1}^{I}$ via an online interaction with the environment, allowing for the exploration of new actions $a_t$ as a function the state $s_t$. In this paper, as illustrated in Fig.~\ref{Online_Offline_Fig}, we assume that the design of policies is carried out on the basis solely of the availability of an offline dataset $\mathcal{D} = \{ (s, a, r, s^{\prime} ) \}$ of transitions $(s, a, r, s^{\prime} )$. Each transition follows the stationary marginal distribution from \eqref{traj_dist}, with policy $\pi(a|s)$ given by the fixed and unknown \emph{behavior policy} $\pi_{\beta}(a | s) = \prod_{i=1}^I \pi_\beta^i(a^i|s)$.
% \begin{equation}
% \label{transtions_distribution}
% (s, a, r, s^{\prime}) \sim P_{\pi_{\beta}}(s) \pi_{\beta}(a | s) R(r_{} | a,s \big) P(s^{\prime}_{} |s,a), 
% \end{equation}
 
 % represents the \emph{behavior policy} used to collect the dataset; $R(r|s,a)$ is the stationary reward distribution; and $P_{\pi_{\beta}}(s)$ is the stationary marginal distribution of state $s$, which is generally a function of the behavioral policy $\pi_{\beta}$.
% The training process of offline RL is quite similar to that of online RL with the main difference being that for offline RL the exploration step is no longer possible and the policy is learned using the offline dataset.

\section{Background}\label{sec:background} %\MS{should this title be about single agent RL ?}
In this section, we present a brief review of distributional RL, as well as of offline RL via CQL for a single agent model~\cite{kumar2020conservative}. This material will be useful to introduce the proposed multi-agent offline DRL solution in the next section.

\subsection{Distributional Reinforcement Learning}

Distributional RL aims at optimizing the agent's policy, $\pi$, while accounting for the inherent aleatoric uncertainty associated with the stochastic environment. To this end, it tracks the return's distribution, allowing the minimization of an arbitrary risk measure, such as the CVaR.

% Thus far, we have defined the return $ Z^{\pi}$ of a whole trajectory.
To elaborate, let us denote the random variable representing the return starting from a given state-action pair $(s, a)$ as $Z^{\pi}(s, a)$. Taking the expectation of the return $Z^{\pi}(s,a)$ over distribution~\eqref{traj_dist} yields the {state-action value function}, also known as \textit{Q-function}, as 
\begin{align}
Q^{\pi}(s,a) = \mathbb{E}\left[Z^{\pi}(s,a) \right].
\label{qfunction}
\end{align}
Classical Q-learning algorithms learn the optimal policy $\pi^*$ by finding the optimal Q-function $Q(s,a)$ as the unique fixed point of the Bellman optimality operator~\cite{bellman1966dynamic}
 % the policy is expressed directly using the Q-function as $\pi(a_t|s_t) = \argmax{} Q(s_t,a_t)$ and
\begin{align}
 Q(s,a)= \mathbb{E}\left[r+ \gamma \:  \max_{a^{\prime}\in \mathcal{A}}Q(s^{\prime},a^{\prime})\right],
% \\
%    & r\sim R(\cdot|s,a), \  s^{\prime}\sim P(\cdot|s,a), \  a^{\prime}\sim \pi(\cdot|s^{\prime}). \nonumber
\label{bell}
\end{align}
with average evaluated with respect to the random variables $(r,s') \sim R(r|s,a)P(s'|s,a)$. The optimal policy $\pi^*$ for the average criterion~\eqref{MARL_Objective_Average} is directly obtained from the optimal Q-function as \begin{equation}
\pi^*(a|s) = \mathbbm{1}\left\{ a = \argmax{a \in \mathcal{A}}\ Q(s,a)\right\},
\label{policy}
\end{equation}
with $ \mathbbm{1}\{\cdot\}$ being the indicator function. 

Similarly, for any risk measure $\rho[\cdot]$, one can define the distributional Bellman optimality operator for the random return $Z^{}(s,a)$ as~\cite{bellemare2017distributional,dabney2017distributional}
\begin{align}
\label{Bellman_opt_Dist_MARL}
   Z(s,a) \overset{D}{=} r + \gamma \: Z\left(s^{\prime}, \argmax{a'\in \mathcal{A}}\ \rho [Z(s^{\prime},a')]\right),
   % & r\sim R(\cdot|s,a), \  s^{\prime}\sim P(\cdot|s,a), \   a^{\prime}\sim \pi(\cdot|s^{\prime}), \nonumber
\end{align}
where equality holds regarding the distribution of the random variables on the left- and right-hand sides, and the random variables $(r,s^{\prime})$ are distributed as in \eqref{bell}. The optimal policy for the general criterion~\eqref{MARL_Objective} can be expressed directly as a function of the optimal $Z(s,a)$ in~\eqref{Bellman_opt_Dist_MARL} as~\cite{bellemare2017distributional,dabney2017distributional}
$$\pi(a|s) = \mathbbm{1}\left\{ a = \argmax{ a  \in \mathcal{A}}\rho [Z(s, a)]\right\}.$$
% where the equality $\overset{D}{=}$ ensures that the left-hand side is distributed according to the same law as the right-hand side.

\emph{Quantile regression DQN} (QR-DQN)~\cite{dabney2017distributional}  estimates the distribution $P_{Z(s,a)}$ of the optimal return $Z(s,a)$ by approximating it via a uniform mixture of Dirac functions centered at $N$ values $\{ \theta_j(s,a)\}_{j=1}^N$, i.e.,
%
% In~\cite{dabney2017distributional}, the authors proposed a quantile regression algorithm, named quantile regression DQN (QR-DQN), to learn the distribution of $Z^{\pi}(s,a)$ at each $(s,a)$ using dynamic programming. Specifically, a uniform mixture of $N$ Diracs is utilized to approximate the distribution $P_{Z^\pi_{}(s,a)}$ of $Z^{\pi}(s,a)$ as follows
\begin{equation}
\label{Diracs}
   \hat{P}_{Z(s,a)} (\theta)=\frac{1}{N} \sum_{j=1}^N \delta_{\theta_j(s,a)}.
\end{equation}
Each value $\theta_j (s,a)$ in~\eqref{Diracs} is an estimate of the quantile $F^{-1}_{{Z(s,a)}}(\hat{\tau}_j)$ of distribution ${P}_{Z(s,a)} $ corresponding to the quantile target $\hat{\tau}_j = (\tau_{j-1}+\tau_j)/2$, with $\tau_j = j/N$ for $1\leq j\leq N$.
Note that $\{ \theta_j(s,a)\}_{j=1}^N$ are estimated via quantile regression, which is achieved by modeling the function mapping $(s,a)$ to the $N$ values $\{ \theta_j(s,a)\}_{j=1}^N$ as a neural network~\cite{dabney2017distributional}, which takes a state as input, and outputs the estimated $\theta_j(s,a)$ for all actions $a\in \mathcal{A}$ . 

The neural network is trained by minimizing the loss
$$
\frac{1}{N^2} \sum_{j=1}^N\sum_{j'=1}^N\zeta_{\hat{\tau}_j}\left(\Delta_{jj^{\prime}}^{} \right),
$$
where $\Delta_{jj'}$ are \emph{the temporal difference (TD) errors} corresponding to the quantile estimates, i.e.,
\begin{align}
\Delta_{jj^{\prime}} = r + \gamma {\theta}_{j^{\prime}}^{}(s^{\prime},a^{\prime }) - \theta_j(s,a),
\label{TDerror_basic}
\end{align}
with $a' = \argmax{a\in \mathcal{A}}\frac{1}{N}\sum_{j'=1}^N{\theta}_{j^{\prime}}^{}(s^{\prime},a^{})$, and $\zeta_{{\tau}}$ is the quantile regression Huber loss defined as
\begin{align}
\label{quantile_loss}
   \zeta_{\tau}^{}(u) &= \begin{cases}-\frac{1}{2} u^2\left|\tau - \mathbbm{1} \{ u < 0 \}\right|, \quad \quad \quad
    \text{if } | u |\leq 1 \\
    \left( | u | - \frac{1}{2}  \right)\left|\tau - \mathbbm{1} \{ u < 0 \}\right|,  \quad 
    \text{otherwise}.
   \end{cases}
   \end{align}

We refer the reader to~\cite{dabney2017distributional} for more details about the theoretical guarantees and practical implementation of QR-DQN.

%\MS{I believe this part is quite clumsy and requires better elaboration / rephrase.}

% Distributional RL shows significant improvements in environments with high uncertainties. This is particularly the case of multi-agent systems where uncertainty is inherently present since each agent is not aware of the other agents' actions~\cite{bellemare2017distributional}.
% Moreover, in environments with risk introduced by taking some actions, distributional RL is extremely useful for designing risk-sensitive policies that optimize certain functions of the return distribution instead of the average return as discussed in the previous section.

\subsection{Conservative Q-Learning}\label{sec:Offline_CQL}

\emph{Conservative Q-learning} (CQL) is a Q-learning variant that addresses epistemic uncertainty in offline RL. Specifically, it tackles the uncertainty arising from the limited available data, which may cause some actions to be OOD due to the lack of exploration. This way, CQL is complementary to QR-DQN, which, instead, targets the inherent aleatoric uncertainty in the stochastic environment. 

To introduce CQL, let us first review conventional DQN~\cite{levine2020offline}, which approximates the solution of the Bellman optimality condition \eqref{bell} by iteratively minimizing the \emph{Bellman loss}
\begin{align}
\label{bellman_error}
    \mathcal{L} (Q, \hat{Q}^{(k)})= \:& \hat{\mathbb{E}} \left[ \left(  r +\gamma \max_{a^{\prime}\in \mathcal{A}} \hat{Q}^{(k)}(s^{\prime},a^{\prime}) 
   - Q(s,a) \right)^2 \right],
\end{align}
% Offline RL suffers from the distributional shift that arises from the out-of-distribution (OOD) actions~\cite{levine2020offline}, namely the actions that are underrepresented in the offline dataset (never or rarely tried by the behavioral policy). To address this problem, a conservative Q-learning (CQL) approach was proposed in~\cite{kumar2020conservative}. Before presenting the CQL approach, we first recall the iterative update to learn the Q-function in \eqref{qfunction} in classical offline DQN~\cite{levine2020offline}
% %\begin{align}
% %\label{Offline_MIN}
% %    \hat{Q} &\leftarrow \argmin{Q} \frac{1}{2} \mathbb{E}_{S_t,A_t,S_{t+1} \sim \mtahcal{D}} \Biggl[ \biggl( \biggl( r(S_t,A_t) +  \\ \nonumber
% %    &\gamma \mathbb{E}_{A_{t+1} \sim \hat{\pi}(A_{t+1} \mid S_{t+1})} \left[ \hat{Q}(S_{t+1},A_{t+1}) \right] \biggr) - Q(S_t,A_t) \biggr)^2 \Biggr],
% %\end{align}
% %which can be described for each individual agent as
% \begin{align}
% \label{Offline_MIN}
%    \hat{Q}^{(k+1)} &\leftarrow \argmin{Q} \: \: \mathcal{L} (Q, \hat{Q}^{(k)}),
% \end{align}
% with $\mathcal{L} (Q, \hat{Q}^{(k)})$ is  defined as
% Training an off-policy RL algorithm, such as offline DQN, without any interaction with the environment suffers from distributional shifts due to the uncorrected OOD actions. Indeed, since the policy is trained to maximize the Q-values, it may overestimate those corresponding to OOD actions. In standard online RL, such overestimation can be corrected by trying the actions and observing the corresponding returns. Unfortunately, this is not possible in offline RL. 
where $\hat{\mathbb{E}}[\cdot]$ is the empirical average over samples $(s,a,r,s')$ from the offline dataset $\mathcal{D}$; $\hat{Q}^{(k)}$ is the current estimate of the optimal Q-function $Q$ at iteration $k$; and the optimization is over function $Q(s,a)$, which is typically modeled as a neural network. The term $ r +\gamma \max_{a^{\prime}\in \mathcal{A}} \hat{Q}^{(k)}(s^{\prime},a^{\prime}) 
   - Q(s,a) $ is also known as the TD-error. The only difference between offline DQN, defined in~\eqref{bellman_error}, and online DQNs lies in the way training data are gathered. For online DQN, the data are collected by interacting with the environment, while learning from the replay buffer, by using stochastic gradient descent (SGD). In contrast, offline DQN has access to a static offline dataset of transitions, or trajectories, that were previously generated by some unknown behavioral policy, and it learns the optimal Q-function by minimizing the Bellman error on the offline dataset over multiple epochs. In this work, we use the term  "offline DQN" to refer to the basic DQN scheme designed for an offline setting. This scheme does not incorporate any modifications intended to mitigate extrapolation errors and overestimation bias that may affect an offline implementation.  Considering the basic offline DQN scheme  will help illustrate the failure of conventional DQN methods in offline settings in the experiments in Section~\ref{sec:results}.

The maximization over the actions in the TD error in (\ref{bellman_error}) may yield over-optimistic return estimates when the Q-function is estimated using offline data. In fact, a large value of the estimated maximum return $\max_{a'\in \mathcal{A}} Q(s,a)$ may be obtained based purely on the randomness in the environment during data collection. This uncertainty could be resolved by collecting additional data. However, this is not possible in an offline setting, and hence one should consider such actions as OOD \cite{levine2020offline,xu2023offline}, and count the resulting uncertainty as part of the epistemic uncertainty.

To account for this issue, the CQL algorithm adds a regularization term to the objective in \eqref{bellman_error} that penalizes excessively large deviations between the maximum estimated return $\max_{a'\in \mathcal{A}} Q(s,a)$, approximated with the differentiable quantity $\log \sum_{\tilde{a}\in \mathcal{A}}  
\exp \bigl( Q(s,\tilde{a}) \bigr)$, and the average value of $Q(s,a)$ in the data set $\mathcal{D}$ as %\MS{what is $\tilde{a}$?}
\begin{align}
 \label{Ind_CQL_Log}
\mathcal{L}_{\text{CQL}}(Q, \hat{Q}^{(k)})=&\: \frac{1}{2}\mathcal{L} (Q, \hat{Q}^{(k)})\\ &+\alpha \hat{\mathbb{E}} \bigg[ \log \sum_{\tilde{a}\in \mathcal{A}}
\exp \bigl( Q(s,\tilde{a}) \bigr) 
    - \ Q(s,a)  \bigg], \nonumber
\end{align}
where $\alpha>0$ is a hyperparameter~\cite{kumar2020conservative}. 

A combination of QR-DQN and CQL was proposed in~\cite{ma2021conservative} for a single-agent setting to address risk-sensitive objectives in offline learning. This approach applies a regularization term as in~\eqref{Ind_CQL_Log} to the distributional Bellman operator~\eqref{Bellman_opt_Dist_MARL}. The next section will introduce an extension of this approach for the multi-agent scenario under study in this paper.

\section{Offline Conservative Distributional MARL with Independent Training}\label{sec:proposed}

%\section{Multi-Agent Offline Conservative Distributional Independent Q-Learning} \MS{This title is too long. I suggest removing "independent" or "offline" from it.}

This section proposes a novel offline conservative distributional independent Q-learning approach for MARL problems. The proposed method combines the benefits of distributional RL and CQL to address the risk-sensitive objective \eqref{MARL_Objective} in multi-agent systems based on offline optimization as in 
Fig.~\ref{Online_Offline_Fig}.  The approaches studied here apply an \emph{independent} Q-learning approach, whereby learning is done separately for each agent. The next section will study more sophisticated methods based on joint training.

\subsection{Multi-Agent Conservative Independent Q-Learning}
\label{subsec:CQL_MARL}
We first present a multi-agent version of CQL, referred to as \emph{multi-agent conservative independent Q-learning} (MA-CIQL), for the offline MARL problem. As in its single-agent version described in the previous section, MA-CIQL addresses the average criterion \eqref{MARL_Objective_Average}, aiming to mitigate the effect of epistemic uncertainty caused by OOD actions. 
 
To this end, each agent $i$ maintains a separable Q-function $Q^i(s,a^i)$, which is updated at each iteration $k$ by approximately minimizing the loss 
% We consider in this work a cooperative multi-agent setting, where all agents cooperate towards the same goal and share the same reward. In addition, we assume decentralized training, where each agent optimizes its policy $\pi^i$ while having full access to the system state. 
% % Hence, each agent has full access to the offline dataset $\mathcal{D}$ except the actions of other agents.
% To this end, the optimal Q-function $Q_i$ corresponding to agent $i$ is obtained using the following iterative update
% \begin{align}
% \label{Ind_CQL_MARL_min}
%     &\hat{Q}_i^{(k+1)} \leftarrow \argmin{Q} \: \: \mathcal{L}_{\text{MA-CQL}}(Q, \hat{Q}_i^{(k)} ),
% \end{align}
% where $\mathcal{L}_{\text{MA-CQL}}(Q, \hat{Q}_i^{(k)} )$ is given by
\begin{align}
\label{Ind_CQL_MARL}
    \mathcal{L}_{\text{MA-CIQL}}&(Q^i, \hat{Q}^{i(k)}) = \frac{1}{2}\mathcal{L} (Q^i, \hat{Q}^{i(k)}) \\ \nonumber&+\alpha \hat{\mathbb{E}} \bigg[\log \bigg( \sum_{\tilde a^i \in \mathcal{A}^i} 
\exp ( Q^i(s,\tilde{a}^i) )\bigg) - Q^i(s, a^i) \bigg],\nonumber
    %+ \frac{1}{2} \mathbb{E}_{s_t,a_t^i,s_{t+1} \sim D} \Biggl[ \biggl( \biggl( r(s_t,a_t^i) +  \\ \nonumber
    %&\gamma \mathbb{E}_{s_{t+1} \sim \hat{\pi}(a_{t+1}^i \mid s_{t+1})} \left[ \hat{Q}(s_{t+1},a_{t+1}^i) \right] \biggr) - Q(s_t,a_t^i) \biggr)^2 \Biggr],
\end{align}
which is the multi-agent version of \eqref{Ind_CQL_Log} over the Q-function $Q^i$, where $\mathcal{L} (Q^i, \hat{Q}^{i(k)})$ is the DQN loss in \eqref{bellman_error} and $\hat{Q}^{i(k)}$ is the estimate of the Q-function of agent $i$ at the $k$-th iteration. Algorithm~\ref{CIQL_MARL} summarizes the MA-CIQL algorithm for offline MARL. Note that the algorithm applies separately to each agent and is thus an example of independent per-agent learning.
% In practice, the Q-function $Q^i$ of agent $i$ is approximated by a neural network, which takes a state $s$ as input and outputs the Q-values $Q^i(s,a^i)$ for all $a^i\in \mathcal{A}^i$. 
% Here, $\hat{\pi}_\beta^i$ is an estimate of the behavioral policy of agent $i$ computed similarly to \eqref{beta_hat}.  Although CQL overcomes the distributional shift that occurs due to the OOD actions, it fails to handle the sources of uncertainties introduced by the other agents and more interestingly from some sources of risk in the environment.

{\LinesNumberedHidden
\begin{algorithm}[!t]
\SetAlgoLined

\textbf{Input:} Discount factor $\gamma$, learning rate $\eta$, conservative penalty constant $\alpha$, number of agents $I$, number of training iterations $K$, number of gradient steps $G$, and offline dataset $\mathcal{D}$

\textbf{Output:} Optimized Q-functions $Q^{i}(s,a^i)$ for $i=1,...,I$

Initialize network parameters %\MS{you need to mention the parameters here by symbols, same for all algorithms} for each agent
%\textbf{Input size:} size of state space. \textbf{Output size:} size of action space.

\For{\text{iteration} $k$ in $\{1$,...,$K$\}}{

\For{\text{gradient step} $g$ in $\{1$,...,$G$\}}{

Sample a batch $\mathcal{B}$ from the dataset $\mathcal{D}$

\For{\text{agent} $i$ in $\{1$,...,$I$\}}{

Estimate the MA-CIQL loss $\mathcal{L}_{\text{MA-CIQL}}$ in~\eqref{Ind_CQL_MARL}

Perform a stochastic gradient step based on the estimated loss %\MS{preferably write the S.G.D equations here. or right it out of the algorithm and refer to it.}

}
}
}
\textbf{Return} $Q^{i}(s,a^i)  = \hat{Q}^{i(K)}(s,a^i)$ for $i=1,...,I$
\caption{Conservative Independent Q-learning for Offline MARL (MA-CIQL)}
\label{CIQL_MARL}  \vspace{0mm}
\end{algorithm}}
 \vspace{-1mm}

\subsection{Multi-Agent Conservative Independent Quantile-Regression}\label{subsec:CQR}
MA-CIQL can only target the average criterion~\eqref{MARL_Objective_Average}, thus not accounting for risk-sensitive objectives that account for the inherent stochasticity of the environment. This section introduces a risk-sensitive Q-learning algorithm for offline MARL to address the more general design objective \eqref{CVaR_Risk} for some risk tolerance level $\xi$. 

 The proposed approach, which we refer to as \emph{multi-agent conservative independent quantile regression} (MA-CIQR), maintains an estimate of the lower tail of the distribution of the return $Z^i(s,a)$, up to the risk tolerance level  $\xi$, for each agent $i$. This is done in a manner similar to~\eqref{Diracs} by using $N$ estimated quantiles, i.e.,
\begin{align}
\label{qr_dist}
     \hat{P}_{Z^i(s,a)}(\theta^i)= \frac{1}{N} \sum_{j=1}^N \delta_{\theta_j^i(s,a)}.
\end{align}
Generalizing~\eqref{Diracs}, however, the quantity $\theta_j^i (s,a)$ is an estimate of the quantile $F^{-1}_{Z^{i}(s,a)}(\hat{\tau}_j)$, with $\hat{\tau}_j = \frac{\tau_{j-1}+\tau_j}{2}$ and $\tau_j = \xi j/N$ for $1\leq j\leq N$. This way, only the quantiles of interest cover the return distribution up to the $\xi$-th quantile.

At each iteration $k$, each agent, $i$, updates the distribution~\eqref{qr_dist} by minimizing a loss function that combines the quantile loss used by QR-DQN and the conservative penalty introduced by CQL. Specifically, the loss function of MA-CIQR is given by
\begin{align}
\label{Ind_CQR}
 &\mathcal{L}_{\text{MA-CIQR}}(\theta^i,\hat{\theta}^{i(k)}) = \frac{1}{2N^2}\hat{\mathbb{E}} \sum_{j=1}^N \sum_{j^{\prime}=1}^N \zeta_{\hat{\tau}_j}\left(\Delta_{jj^{\prime}}^{i(k)} \right) \\ &
\quad\quad +\alpha  \hat{\mathbb{E}}\Biggl[ \frac{1}{N} \sum_{j=1}^N \Biggl[\log \sum_{\tilde{a}^i \in \mathcal{A}^i} 
\exp \bigl( \theta_j^i(s,\tilde{a}^i) \bigr)
    -  \theta_j^i(s,a^i)  \Biggr] \Biggl], \nonumber 
\end{align}
where $\zeta_{\tau}^{}(u)$ is the quantile regression Huber loss defined in \eqref{quantile_loss}
% \begin{align}
% \label{quantile loss}
%    \zeta_{\tau}^{}(u) &= \begin{cases}-\frac{1}{2} u^2\left|\tau - \mathbbm{1} \{ u < 0 \}\right|, \quad \quad \quad
%     \text{if } | u |\leq 1 \\
%     \left( | u | - \frac{1}{2}  \right)\left|\tau - \mathbbm{1} \{ u < 0 \}\right|,  \quad 
%     \text{otherwise};
%    \end{cases}
%    \end{align}
and $\Delta^{i(k)}_{j j^{\prime}}$ are the TD errors evaluated with the quantile estimates as
\begin{align}
\Delta_{jj^{\prime}}^{i(k)} = r + \gamma \hat{\theta}_{j^{\prime}}^{i(k)}(s^{\prime},a^{\prime i}) - \theta_j^i(s,a^i),
\label{TDerror}
\end{align}
where $a^{\prime i} = \argmax{a^i\in\mathcal{A}^i} \frac{1}{N}\sum_{j^{\prime}=1}^N \hat{\theta}_{j^{\prime}}^{i(k)}(s^{\prime},a^i)$. 
% where $\kappa$ is a fixed threshold and $\mathcal{L}_\kappa(u)$ is the Huber loss
%    \begin{align}
%    \label{huber_loss}
%    \mathcal{L}_\kappa(u)&= \begin{cases}
%     - \frac{1}{2} u^2, & \quad
%     \text{if } \mid u \mid \leq \kappa \\
%     \kappa \left( \mid u \mid - \frac{1}{2} \kappa \right) & \quad 
%     \text{otherwise}.
%     \end{cases}
% \end{align}
Note that the TD error $\Delta_{jj^{\prime}}^{i(k)}$ is obtained by using the $j'$-th quantile of the current $k$-th iteration to estimate the return as $r + \gamma \hat{\theta}_{j^{\prime}}^{i(k)}(s^{\prime},a^{\prime i})$, while considering the $j$-th quantile $\theta_j^i(s,a^i)$ as the quantity to be optimized. 

The corresponding optimized policy is finally obtained as 
\begin{equation}\label{eq:polours}
\pi^i(a^{i}|s) = \mathbbm{1} \left\{  a^i = \argmax{ a^i \in \mathcal{A}^i} \frac{1}{N}\sum_{j=1}^N {\theta}_{j}^{i}(s^{\prime},  a^i)\right\}. 
\end{equation} By (\ref{CVaR_obj}), the objective in (\ref{eq:polours}) is an estimate of the CVaR at the risk tolerance level $\xi$. 
The pseudocode of the MA-CIQR algorithm is provided in Algorithm~\ref{CIQR_MARL}. As for MA-CIQL, MA-CIQR applies separately across all agents.

{\LinesNumberedHidden
\begin{algorithm}[!t]
% \SetAlgoLined

    \textbf{Input:} Discount factor $\gamma$, learning rate $\eta$, number of quantiles $N$, conservative penalty constant $\alpha$, number of agents $I$, number of training iterations $K$, number of gradient steps $G$, offline dataset $\mathcal{D}$, and CVaR parameter $\xi$

    \textbf{Output:} Optimized quantile estimates $\{ {\theta}_j^i(s,a^i)\}_{j=1}^N$ for all $i=1,...,I$

Define $\tau_i =  \xi i/N, \ i=1,...,N$\\

Initialize network parameters for each agent
%\textbf{Input size:} size of state space. \textbf{Output size:} size of action space $\times$ $N$.

\For{\text{iteration} $k$ in $\{1$,...,$K$\}}{

\For{\text{gradient step} $g$ in $\{1$,...,$G$\}}{

Sample a batch $\mathcal{B}$ from the dataset $\mathcal{D}$

\For{\text{agent} $i$ in $\{1$,...,$I$\}}{

\For{$j$ in $\{1$,...,$N$\}}{

\For{$j^{\prime}$ in $\{1$,...,$N$\}}{

%Calculate $\mathcal{L}_\kappa( \delta_{jk}^i)$ using~\eqref{Distr_Loss_MARL}.

Calculate TD errors $\Delta_{j j^{\prime}}^{i(k)}$ 
using \eqref{TDerror}
}
% Average the conservative loss over the $N$ quantiles: $\log \sum_{a^i} 
% \exp \bigl( \theta_j^i(s_t,a_t^i) \bigr) - \mathbb{E}_{a_t^i \sim \hat{\pi}_{\beta}(a^i \mid s)} \bigl[ \theta_j^i(s_t,a_t^i) \bigl]$

}

Estimate the MA-CIQR loss $\mathcal{L}_{\text{MA-CIQR}}$ in ~\eqref{Ind_CQR}

Perform a stochastic gradient step based on the estimated loss %\MS{ I also suppose it is good to mention the S.G.D equation somewhere or some reference for it. same for all algorithms }
% Update network parameters: $\tilde{\phi}^i\coloneqq \tilde{\phi}^i - \eta \nabla_{\tilde{\phi}^i} \mathcal{L}_{\text{MA-CQR}}$

}
}
}
\textbf{Return} $ \{ {\theta}_j^i(s,a^i)\}_{j=1}^N =  \{ \hat{\theta}_j^{i(K)}(s,a^i)\}_{j=1}^N  $ for all $i=1,...,I$
\caption{Conservative Independent Quantile Regression for Offline MARL (MA-CIQR) %\MS{would be much better to add step numbers on the left. Same for all agorithms}
}
\label{CIQR_MARL}  \vspace{0mm}
\end{algorithm}}
 \vspace{-1mm}

\section{Offline Conservative Distributional MARL with Centralized Training}\label{sec:proposed_Cent}

%\section{Multi-agent Offline Conservative Distributional Q-Learning with Centralized Training} \MS{Shall we make this title shorter?}

The independent learning strategies studied in the previous section may fail to yield coherent policies across different agents. This section addresses this issue by introducing joint / centralized methods based on \emph{value decomposition}~\cite{sunehag2017value}. %Specifically, we adopt the CTDE framework, which supports the optimization of the agents' policies based on a global loss, along with the decentralized execution of the optimized policies. 

\subsection{Multi-Agent Conservative Centralized Q-Learning}

{\LinesNumberedHidden
\begin{algorithm}[!t]
\SetAlgoLined

\textbf{Input:} Discount factor $\gamma$, learning rate $\eta$, conservative penalty constant $\alpha$, number of agents $I$, number of training iterations $K$, number of gradient steps $G$, and offline dataset $\mathcal{D}$

\textbf{Output:} Optimized Q-functions $Q^{i}(s,a^i)$ for $i=1,...,I$

Initialize network parameters for each agent
%\textbf{Input size:} size of state space. \textbf{Output size:} size of action space.

\For{\text{iteration} $k$ in $\{1$,...,$K$\}}{

\For{\text{gradient step} $g$ in $\{1$,...,$G$\}}{

Sample a batch $\mathcal{B}$ from the dataset $\mathcal{D}$

Estimate the MA-CCQL loss $\mathcal{L}_{\text{MA-CCQL}}$ in~\eqref{Ind_CQL_CTDE}

Perform a stochastic gradient step to update the network parameters of each agent

}
}
\textbf{Return} $\tilde{Q}^{i}(s,a^i)  = \hat{Q}^{i(K)}(s,a^i)$, for $i=1,...,I$
\caption{Conservative Centralized Q-learning for Offline MARL (MA-CCQL) }
\label{CCQL_MARL}  \vspace{0mm}
\end{algorithm}}
 \vspace{-1mm}

% Can we come up with an improved version that applies learning across the agents?
With value decomposition, it is assumed that the global Q-function can be written as~\cite{sunehag2017value}
\begin{align}
\label{value_dec_eq}
   Q(s,a) = \sum_{i=1}^I \tilde{Q}^i(s,a^i),
\end{align}
where the function $\tilde{Q}^i(s,a^i)$ indicates the contribution of the $i$-th agent to the overall Q-function. For conventional DQN, the Bellman loss~\eqref{bellman_error} is minimized over the functions $\{\tilde{Q}^i(s,a^i)\}_{i=1}^I$. This problem corresponds to the minimization of the global loss
\begin{align}
\label{bellman_error_global}
    \mathcal{L} (\{\tilde{Q}^i\}_{i=1}^I, \{\hat{Q}^{i(k)}\}_{i=1}^I) = & \hat{\mathbb{E}} \Bigg[ \Bigg( r +\gamma \sum_{i=1}^I\max_{\tilde{a}^{i}\in \mathcal{A}^i} \hat{Q}^{i(k)}(s^{\prime},a^{i})\nonumber\\
    &-\sum_{i=1}^I \tilde{Q}^i(s,a^i)\Bigg)^2 \Bigg],
\end{align}
where $\hat{Q}^{i(k)}$ is the current estimate of the contribution of agent $i$.
%\begin{align}
%\label{bellman_error_global}
%    &\mathcal{L} (Q, \hat{Q}^{(k)})= \:\frac{1}{2} \hat{\mathbb{E}} \left[ \left(  r +\gamma \max_{a^{\prime}\in \mathcal{A}} \hat{Q}^{(k)}(s^{\prime},a^{\prime}) 
%   - Q(s,a) \right)^2 \right],\nonumber\\
%  &=\frac{1}{2} \hat{\mathbb{E}} \left[ \left(  r +\gamma \sum_{i=1}^I\max_{a^{\prime i}\in \mathcal{A}^i}  \hat{\tilde Q}^{i(k)}(s^{\prime},a^{\prime i}) 
%   -\sum_{i=1}^I \tilde{Q}^i(s,a^i)\right)^2 \right],
%\end{align}
In practice, every function $\tilde{Q}^i(s,a^i)$ is approximated using a neural network. Furthermore, the policy of each agent is obtained from the optimized function $\tilde{Q}^i(s,a^i)$ as
\begin{equation}
\label{Policy_opt_ind}
\pi^i(a^i|s) = \mathbbm{1}\left\{ a^i = \argmax{a^i \in \mathcal{A}^i}\ \tilde{Q}^i(s,a^i)\right\}.
\end{equation}

The same approach can be adopted to enhance MA-CIQL by using~\eqref{value_dec_eq}  in the loss~\eqref{Ind_CQL_MARL}. This yields the loss 
\begin{align}
\label{Ind_CQL_CTDE}
  &\mathcal{L}_{\text{MA-CCQL}} (\{\tilde{Q}^i\}_{i=1}^I, \{\hat{Q}^i\}_{i=1}^I) =  \frac{1}{2} \mathcal{L} (\{\tilde{Q}^i\}_{i=1}^I, \{\hat{Q}^i\}_{i=1}^I) \nonumber \\ & +\alpha \hat{\mathbb{E}} \sum_{i=1}^I\bigg[\log \bigg( \sum_{\tilde a^i \in \mathcal{A}^i} 
\exp ( \tilde{Q}^i(s,\tilde{a}^i) )\bigg)  
     - \tilde{Q}^i(s, a^i) \bigg],
    %+ \frac{1}{2} \mathbb{E}_{s_t,a_t^i,s_{t+1} \sim D} \Biggl[ \biggl( \biggl( r(s_t,a_t^i) +  \\ \nonumber
    %&\gamma \mathbb{E}_{s_{t+1} \sim \hat{\pi}(a_{t+1}^i \mid s_{t+1})} \left[ \hat{Q}(s_{t+1},a_{t+1}^i) \right] \biggr) - Q(s_t,a_t^i) \biggr)^2 \Biggr],
\end{align}
with $\mathcal{L} (\{\tilde{Q}^i\}_{i=1}^I, \{\hat{Q}^i\}_{i=1}^I)$ defined in~\eqref{bellman_error_global}. The obtained scheme, whose steps are detailed in Algorithm \ref{CCQL_MARL}, is referred to as \emph{multi-agent conservative centralized Q-learning} (MA-CCQL). The optimized policy is given in~\eqref{Policy_opt_ind}.

%\begin{remark}
%There are two options to define the conservative penalty here. The first option, adopted in \eqref{Ind_CQL_CTDE}, consists of having a penalty function for each agent and summing them. The second option is to define a joint penalty function based on the global Q-function \eqref{value_dec_eq}. The latter would be better theoretically as it reflects a penalty on the global $Q(s,a)$, but the former is more efficient from a practical point of view. In fact, estimating the term $\log \sum_{a} \exp$ from the offline dataset considering all possible global actions $a\in\mathcal{A}$ is not efficient in practice. We have tested both options and we found that the first one is more efficient.
%\label{rem}
%\end{remark}

{\LinesNumberedHidden
\begin{algorithm}[!t]
% \SetAlgoLined

    \textbf{Input:} Discount factor $\gamma$, learning rate $\eta$, number of quantiles $N$, conservative penalty constant $\alpha$, number of agents $I$, number of training iterations $K$, number of gradient steps $G$, offline dataset $\mathcal{D}$, and CVaR parameter $\xi$

    \textbf{Output:} Optimized functions $\{ \tilde{\theta}_j^i(s,a^i)\}_{j=1}^N$ for all $i=1,...,I$

Define $\tau_i =  \xi i/N, \ i=1,...,N$\\

Initialize network parameters for each agent
%\textbf{Input size:} size of state space. \textbf{Output size:} size of action space $\times$ $N$.

\For{\text{iteration} $k$ in $\{1$,...,$K$\}}{

\For{\text{gradient step} $g$ in $\{1$,...,$G$\}}{

Sample a batch $\mathcal{B}$ from the dataset $\mathcal{D}$

\For{$j$ in $\{1$,...,$N$\}}{

\For{$j^{\prime}$ in $\{1$,...,$N$\}}{

% \For{\text{agent} $i$ in $\{1$,...,$I$\}}{

% Estimate the individual distributional Q-functions and Q-targets for each agent

% Estimate the individual conservative loss for each agent

% }

Calculate global TD error $\Delta_{j j^{\prime}}^{(k)}$ using~\eqref{TDerror_global_CTDE}

% Average the conservative loss over the $N$ quantiles: $\log \sum_{a^i} 
% \exp \bigl( \theta_j^i(s_t,a_t^i) \bigr) - \mathbb{E}_{a_t^i \sim \hat{\pi}_{\beta}(a^i \mid s)} \bigl[ \theta_j^i(s_t,a_t^i) \bigl]$

}
}
Estimate the MA-CCQR loss $\mathcal{L}_{\text{MA-CCQR}}$ in~\eqref{Ind_CQR_CTDE}

Perform a stochastic gradient step to update the
network parameters of each agent
% Update network parameters: $\tilde{\phi}^i\coloneqq \tilde{\phi}^i - \eta \nabla_{\tilde{\phi}^i} \mathcal{L}_{\text{MA-CQR}}$

}
}
\textbf{Return} $\{ \hat{\theta}_j^{i(K)}(s,a^i)\}_{j=1}^N$, for $i=1,...,I$
\caption{Conservative Centralized Quantile Regression for Offline MARL (MA-CCQR)}
\label{CQR_CTDE}  \vspace{0mm}
\end{algorithm}}
 \vspace{-1mm}

\subsection{Multi-Agent Conservative Centralized Quantile-Regression}

The joint training approach based on the value decomposition~\eqref{value_dec_eq} can also be applied to MA-CIQR to obtain a centralized training version referred to as \emph{multi-agent conservative centralized quantile regression} (MA-CCQR). 

To this end, we first recall that the lower tail of the distribution of $Z(s,a)$ is approximated by MA-CIQR as in \eqref{qr_dist}
%\begin{align}
%\label{qr_dist_CTDE}
%     \hat{P}_{{Z}(s,a)}(\tilde{\theta})= \frac{1}{N} \sum_{j=1}^N \delta_{{\theta}_j(s,a)},
%\end{align}
using the estimates of the quantiles $F^{-1}_{Z(s,a)}(\hat{\tau}_j)$, with $\hat{\tau}_j = \frac{\tau_{j-1}+\tau_j}{2}$ and $\tau_j = \xi j/N$ for $1\leq j\leq N$. To jointly optimize the agents' policies, we decompose each quantile ${\theta}_j(s,a)$ as
\begin{equation}\label{dist_value_dec}
{\theta}_j(s,a) = \sum_{i=1}^I \tilde{\theta}^i_j(s,a^i), 
\end{equation}
where $\tilde{\theta}^i_j(s,a^i) $ represents the contribution of agent $i$. The functions $\{\{\tilde{\theta}^i_j(s,a^i)\}_{i=1}^I\}_{j=1}^N$ are jointly optimized using a loss obtained by plugging the decomposition~\eqref{dist_value_dec} into~\eqref{Ind_CQR} to obtain
\begin{align}
\label{Ind_CQR_CTDE}
 \mathcal{L}&_{\text{MA-CCQR}}(\{\{\tilde{\theta}^i_j\}_{i=1}^I\}_{j=1}^N,\{\{\hat{\theta}^{i(k)}_j\}_{i=1}^I\}_{j=1}^N) = \nonumber\\
 &\: \: \: \frac{1}{2N^2}\hat{\mathbb{E}} \sum_{j=1}^N \sum_{j^{\prime}=1}^N \zeta_{\hat{\tau}_j}\left(\Delta_{jj^{\prime}}^{(k)} \right) \\ &\: \: \: + \alpha  \hat{\mathbb{E}} \sum_{i=1}^I \Biggl[ \frac{1}{N} \sum_{j=1}^N \Biggl[\log \sum_{\tilde{a} \in \mathcal{A}} 
\exp \left(\tilde{\theta}_j^i(s,\tilde{a}^i) \right)
    - \tilde{\theta}_j^i(s,a^i)  \Biggr] \Biggl], \nonumber 
\end{align}
where $\{\{\hat{\theta}^{i(k)}_j\}_{i=1}^I\}_{j=1}^N$ represents the current estimate of the contribution of agent $i$ and $\Delta_{jj^{\prime}}^{(k)}$ is given by
\begin{align}
\Delta_{jj^{\prime}}^{(k)} = r + \gamma \sum_{i=1}^I \hat{\theta}_{j^{\prime}}^{i(k)}(s^{\prime},a^{\prime i}) - \sum_{i=1}^I\tilde{\theta}_j^i(s,a^i),
\label{TDerror_global_CTDE}
\end{align}
with $a^{\prime i} = \argmax{a^i\in\mathcal{A}^i} \frac{1}{N}\sum_{j^{\prime}=1}^N \hat{\theta}_{j^{\prime}}^{i(k)}(s^{\prime},a^i)$. %Note that as in MA-CCQL, we consider here individual conservative penalty functions instead of a global one for the same reasons explained in Remark \ref{rem}. 
The individual policies of the agent are finally obtained as 
$$
\pi^i(a^i|s) =  \mathbbm{1}\left\{ a^i = \argmax{a^i \in \mathcal{A}^i}\  \frac{1}{N}\sum_{j^{}=1}^N {\tilde{\theta}}_{j^{}}^{i}(s^{},a^i)\right\}.
$$
For each agent, the function that maps $(s,a^i)$ to the $N$ values $\{ \tilde{\theta}_j^i(s,a)\}_{j=1}^N$ is modeled as a neural network and the steps of the MA-CCQR scheme are provided in Algorithm \ref{CQR_CTDE}.

\section{Application: Trajectory Learning in UAV Networks}\label{sec:results} %\MS{I suggest "Application: Trajectory Learning (or optimization) in UAV Networks".}
In this section, we consider the application of offline MARL to the trajectory optimization problem in UAV networks. Following~\cite{9815722}, as illustrated in Fig. \ref{System_Model_Fig}, we consider multiple UAVs acting as BSs to receive uplink updates from limited-power sensors. %The design objective is to minimize power expenditure while minimizing the age of information (AoI) for data retrieved from the sensors. \MS{AoI should be defined and discussed here.}

% First, we present the UAV network environment and then the offline dataset collection. Then, we show the simulation results of the proposed method compared to the baseline schemes.
\subsection{Problem Definition and Performance Metrics}\label{sec:sysmodel}
\begin{figure}[t!]
    \centering    \includegraphics[width=1\columnwidth,trim={0 0 0 0},clip]{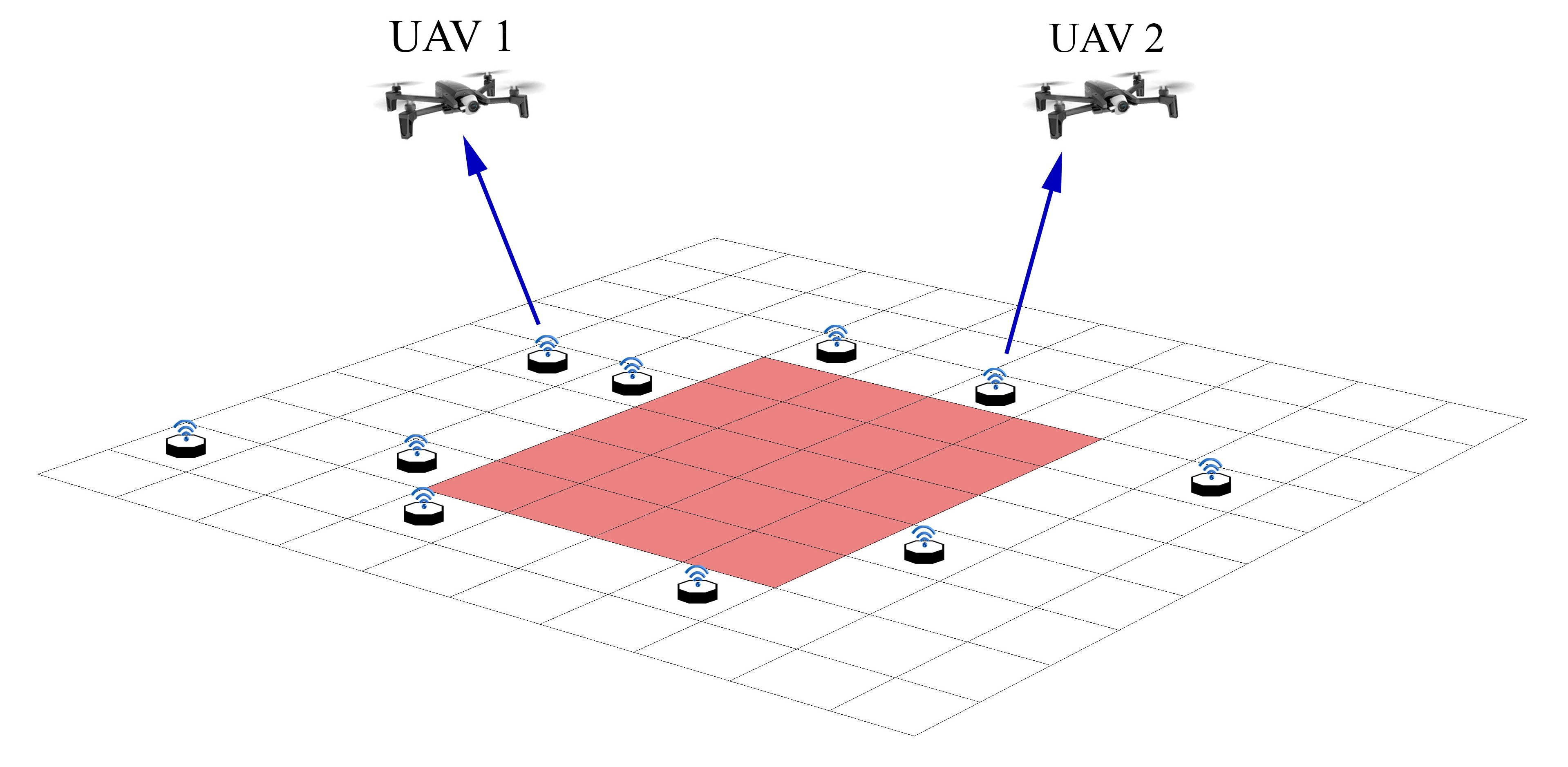} \vspace{2mm}
    \caption{Multiple UAVs serve limited-power sensors to minimize power expenditure while also minimizing the age of information for data retrieval from the sensors. The environment is characterized by a {risk} region for navigation of the UAVs in the middle of the grid world (colored area).}
    \vspace{0mm}
    \label{System_Model_Fig}
\end{figure}
Consider a grid world, as shown in Fig.~\ref{System_Model_Fig}, where each cell is a square of length $L_c$. The system comprises a set $\mathcal{M}$ of $M$ uplink IoT devices deployed uniformly in the grid world. The devices report their observations to $I$ fixed-velocity rotary-wing UAVs flying at height $h$ and starting from positions selected randomly on the grid. The grid world contains normal cells, represented as white squares, and a \emph{risk region} of special cells, colored in the figure. The risk region can be an area with a high probability of UAV collision and/or locations with a high chance of signal blockages. The current position at each time $t$ of each UAV $i$ is projected on the plane as coordinates $\left(x^i_t,y^i_t \right)$. The goal is to determine trajectories for the UAVs on the grid that jointly minimize the AoI and the transmission powers across all the IoT devices. 

The AoI measures the freshness of the information collected by the UAVs from the devices~\cite{9815722}. For each device $m$, the AoI is defined as the time elapsed since the last time data from the device was collected by a UAV~\cite{9845353,AoI}. Accordingly, the AoI of device $m$ at time $t$ is updated as follows
\begin{equation}
\label{AoI_update}
	A_t^m =
	\begin{cases}
		1, & \quad \text{if} \ V_t^m=1, \\
		\text{min}\{A_{\rm max},A_{t-1}^m + 1\}, & \quad \text{otherwise};
	\end{cases}
\end{equation}
where $A_{\rm max}$ is the maximum AoI, and $V_t^m = 1$ indicates that device $m$ is served by a UAV at time step $t$. The maximum value $A_{\rm max}$ determines the maximum penalty assigned to the UAVs for not collecting data from a device at any given time.

For the sake of demonstrating the idea, we assume line-of-sight (LoS) communication links and write the channel gain between agent $i$ and device $m$ at time step $t$ as
\begin{equation}
\label{ch_gain_d}
    g_{t}^{i, m} = \frac{g_0}{h^2+(L_{t}^{i, m})^2},
\end{equation}
where $g_0$ is the channel gain at a reference distance of $1$ m and $L_{t}^{i,m}$ is the distance between UAV $i$ and device $m$ at time $t$. Using the standard Shannon capacity formula, for device $m$ to communicate to UAV $i$ at time step $t$, the transmission power must be set to \cite{10293964}
\begin{align}
\label{Tx_P_1}
P_t^{i, m} &= \frac{\left(2^{\frac{E}{B}}\!-1\right) \sigma^2}{g_{t}^{i, m}}, 
    % &= 
    % \left(2^{\frac{E}{B}}-1\right)\frac{\sigma^2}{g_0}\:\left( h^2+L_{t}(i_m, m)^2\right),\label{Tx_P_2}
\end{align}
where $E$ is the size of the transmitted packet, $B$ is the bandwidth, and $\sigma^2$ is the noise power.

%\begin{figure}[t!]
%    \centering
%    \subfloat[Risk-neutral MARL\label{Traj_CQL}]{\includegraphics[width=0.45\textwidth,trim={0 0 0 0},clip]{Figures/Risk_Neutral.pdf}}
%    \hskip -1.5ex
%    \subfloat[Risk-sensitive MARL\label{Traj_CVAR}]{\includegraphics[width=0.45\textwidth,trim={0 0 0 0},clip]{Figures/Risk_Sensitive.pdf}}
%    \caption{A comparison between the trajectories of multiple UAVs using risk-neutral and risk-sensitive algorithms. A risk-neutral algorithm may not take into account avoiding the risk sources in the environment, whereas a risk-sensitive algorithm avoids entering the risk region.}
%    \label{Traj_Fig} \vspace{-2mm}
%\end{figure}

If all the UAVs are outside the risk region, the {reward function} is given deterministically as a weighted combination of the sums of AoI and powers across all agents
\begin{equation}
\label{Reward_fn}
r_t = - \frac{1}{M}\sum_{m=1}^M A_t^m - \lambda \sum_{m=1}^M P_t^{i_m, m},
\end{equation}
where $\lambda>0$ is a parameter that controls the desired trade-off between  AoI and power consumption. In contrast, if any of the UAVs is within the risk region, with probability $p_\mathrm{risk}$, the reward is given by (\ref{Reward_fn}) with the addition of a penalty value $\mathrm{P_\mathrm{risk}}>0$, while it is equal to (\ref{Reward_fn}) otherwise. For instance, if the risk region corresponds to an area with a high probability of signal blockages, the penalty $\mathrm{P_\mathrm{risk}}$ may be chosen to be proportional to the amount of power needed to resend the packets lost due to blockages of the communication links between the UAVs and the sensors. That said, it is emphasized that the proposed model is general and that the specific application scenario would practically dictate the choice of the penalty $\mathrm{P_\mathrm{risk}}$. 

% penalized Furthermore,  to take into account the uncertainty in the environment, we consider a risk region in the middle of the grid world, which can be interpreted as a highly secured area or an area with many blockages that affect the communication and the battery of the devices. Whenever an agent passes through this region at timestep $t$, there is a small probability $p_{risk}$ of receiving a high penalty $\mathcal{P}$ included in the reward function $r_t$ as follows
% \begin{align}
% \label{Reward_Risk}
% 	r_t =
% 	\begin{cases}
% 		- \frac{\sum_{m=1}^M A_t^m}{M} - \lambda \sum_{m=1}^M P_t^{i_m, m} - \mathcal{P}, & \text{w. p. } p_{risk} \\
% 		- \frac{\sum_{m=1}^M A_t^m}{M} - \lambda \sum_{m=1}^M P_t^{i_m, m}, & \text{w. p. } 1-p_{risk}
% 	\end{cases}
% \end{align}
% % where $\mathcal{P}$ is a high penalty constant and $"\text{if risk}"$ is \textbf{True} with probability $p_{risk}$ whenever the agent enters the risk region.
% Specifically, the reward function is computed using \eqref{Reward_fn} if all agents are outside the risk region and using \eqref{Reward_Risk} if at least one of the agents is inside the risk region.
 
To complete the setting description, we define state and actions as follows. The global state of the system at each time step $t$ is the collection of the UAVs' positions and the individual AoI of the devices, i.e., $s_t = \left[ x_t^1, y_t^1, \cdots, x_t^I, y_t^I, A_t^1, A_t^2, \cdots, A_t^M \right]$. %Each agent has his own observations $o_t^u = \big[ x_t^u, y_t^u,\\ A_t(1), \cdots, A_t(M) \big]$. Herein, we assume an MMDP, where $S(t) = o_t^1 = \cdots = o_t^U$. 
At each time $t$, the action $a_t^i = [ w_t^i, d_t^i]$ of each  UAV $i$ includes the direction $w^i_t\in \{\text{north}, \text{south}, \text{east}, \text{west}, \text{hover} \}$, where ``{hover}'' represents the decision of staying in the same cell, while the other actions move the UAV by one cell in the given direction. 
It also includes the identity $d_t^i \in \mathcal{M}\cup\{0\}$ of the device served at time $t$, with $d_t^i=0$ indicating that no device is served by UAV $i$.

 % Fig.~\ref{Traj_Fig} depicts a $2-$D visualization of the trajectories of two UAVs in the environment corresponding to two different policies, risk-sensitive and risk-neutral. We can see from the figure that the risk-sensitive agent is able to avoid the risk region.

 % Risk-neutral objectives, as in~\eqref{MARL_Objective_Average}, maximizes the average return without considering risk outcomes. On the other hand, risk-sensitive objectives, as CVaR in~\eqref{CVaR_obj}, accounts on the extreme scenarios and thus, avoids the risk region.

% For the performance metrics, we consider the return convergence over epochs during training as an initial indicator to the learning of each scheme. For the first set of simulations, we focus on the mean return to be the objective of the system. We compare all the schemes in terms of the least feasible average AoI and total power consumption of the system. Finally, we consider evaluating the most extreme scenarios with the worst returns.

\subsection{Implementation and Dataset Collection}

% \begin{table}[t!]
% \centering
% \caption{The simulation hyperparameters of the UAV network and the proposed CQR algorithm.}
% \label{UAV_Parameters}
% \begin{tabular}{cc|cc}
% \toprule
% \textbf{Parameter}                                    & \textbf{Value} & \textbf{Parameter}                                    & \textbf{Value} \\ \midrule

% $g_0$ & $30$ dB & $\sigma^2$ & $-100$ dBm \\
% $B$ & $1$ MHz & $E$ & $5$ Mb \\
% $h$ & $100$ m & $L_c$ & $100$ m \\
% $\lambda$ & $500$ & $\mathcal{P}$ & $\frac{\lambda}{4}$ \\https://www.overleaf.com/project/653f99d978816403a88fd7bd
% $\alpha$ & $1$ & $\gamma$ & $0.99$ \\
% $A_{max}$ & 100 & $\xi$ & $0.15$ \\
% $N$ & $8$ &  \\
% $p_{risk}$ & 0.1 &  Learning rate & $10^{-4}$\\
% Hidden layers & $(256,256)$ & Optimizer & Adam \\
% Iterations $K$ & $150$ & Batch size & $128$ \\

% \bottomrule
% \end{tabular} 
% \end{table}

\begin{table}[t!]
\centering
\caption{Simulation parameters and hyperparameters}
\label{UAV_Parameters}
\begin{tabular}{cc|cc}
\toprule
\textbf{Parameter}                                    & \textbf{Value} & \textbf{Parameter}                                    & \textbf{Value} \\ \midrule
\midrule

$g_0$ & $30$ dB & $\alpha $ & $1$\\
$B$ & $1$ MHz & $\gamma$ & $0.99$\\
$h$ & $100$ m & $\xi$ & $0.15$\\
$E$ & $5$ Mb & $\sigma^2$ & $-100$ dBm\\
$\gamma$ & $0.99$ & $A_{\max}$ & 100 \\
Batch size & $128$ & $\lambda$ & $500$\\ 
Iterations $K$ & $150$ & $p_{\mathrm{risk}}$ & 0.1 \\
$L_c$ & $100$ m & Optimizer & Adam \\

\bottomrule
\end{tabular} 
\end{table}

%Learning rate & $10^{-4}$
%$\mathrm{P_{risk}}$& ${\lambda}/{4}$

We consider a $10 \times 10$ grid world with $I=2$ UAVs serving $M=10$ limited-power sensors and a $5 \times 4$ risk region in the middle of the grid world as illustrated in Fig.~\ref{System_Model_Fig}. We use a fully connected neural network with two hidden layers of size $256$ and ReLU activation functions to represent the Q-function and the quantiles. The experiments are implemented using Pytorch on a single NVIDIA Tesla V100 GPU. Table~\ref{UAV_Parameters} shows the UAV network parameters and the proposed schemes' hyperparameters. We compare the proposed method MA-CQR to baseline offline MARL schemes, namely multi-agent deep Q-network (MA-DQN) \cite{tampuu2017multiagent}, MA-CQL (see Sec.\ref{subsec:CQL_MARL}), and multi-agent quantile regression DQN (MA-QR-DQN). MA-DQN corresponds to MA-CQL when no conservative penalty for OOD is applied, i.e., $\alpha = 0$ in \eqref{Ind_CQL_MARL}, whereas MA-QR-DQN corresponds to MA-CQR when $\alpha = 0$ and $\xi = 1$. Both the independent and centralized training frameworks apply to MA-DQN, yielding multi-agent deep independent Q-network (MA-DIQN) and multi-agent deep centralized Q-network (MA-DCQN), and to MA-QR-DQN, yielding multi-agent quantile regression deep independent Q-network (MA-QR-DIQN) and multi-agent quantile regression deep centralized Q-network (MA-QR-DCQN).

For the proposed MA-CQR, we consider two settings for the risk tolerance level $\xi$, namely $\xi=1$ and $\xi=0.15$, with the former corresponding to a risk-neutral design. We refer to the former as MA-CQR and the latter as MA-CQR-CVaR. For all distributional RL schemes (MA-QR-DQN and MA-CQR in all its variants), the learning rate is set to $10^{-5}$, while for all other schemes, we use a learning rate of $10^{-4}$.

The offline dataset $\mathcal{D}$ is collected using online independent DQN agents. In particular, we train the UAVs using an online MA-DQN algorithm until convergence and use $6\%$  and $16\%$ of the total number of transitions from the observed experience as the offline datasets\footnote{The code and datasets are available at \url{https://github.com/Eslam211/Conservative-and-Distributional-MARL}}.

\begin{figure}[t!]
    \centering    \includegraphics[width=1\columnwidth,trim={0 0 0 0},clip]{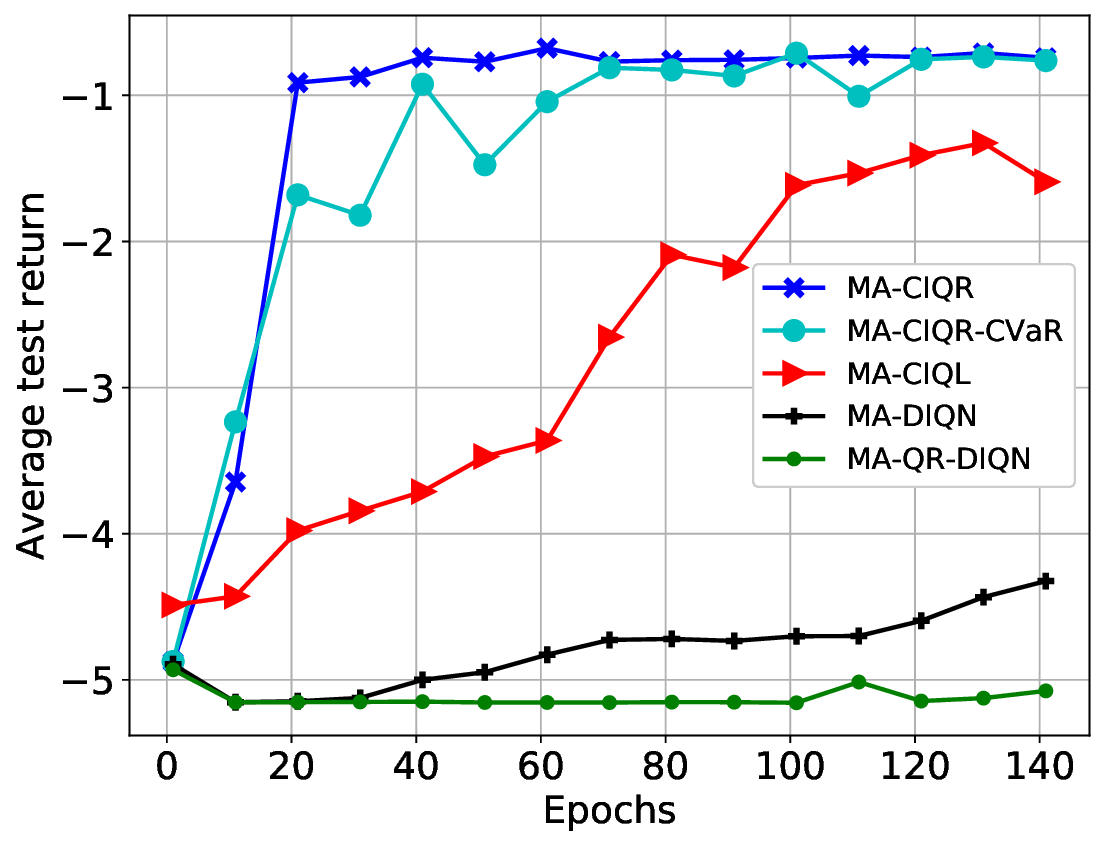} \vspace{2mm}
    \caption{Average test return as a function of the number of training epochs using $\mathrm{P_{risk}}=100$ and $16 \%$ offline dataset for a system of $2$ UAVs serving $10$ sensors. The return is averaged over $100$ test episodes at the end of each training epoch and shown upon division by $1000$.
    % We notice that DQN and QR-DQN fails to converge, whereas CQR provides a smooth and stable learning convergence after $30$ epochs.
    } 
    \vspace{0mm}
    \label{Convergence_Fig}
\end{figure}

%\begin{figure}[t!]
%    \centering
%    \subfloat[Dataset of $5000$ rows]{\includegraphics[width=0.44\textwidth,trim={0.1 0 0 0},clip]{Figures/Conv_2_5000.eps} \label{Conv_2_5000}} 
%    \hskip -2.28ex
%    \subfloat[Dataset of $2000$ rows]{\includegraphics[width=0.44\textwidth,trim={0.1 0 0 0},clip]{Figures/Conv_2_2000.eps} \label{Conv_2_2000}} 
    %\hskip -2.28ex
%    \caption{Average test return as a function of the number of training epochs at the DT for an an environment of $2$ agents and $10$ sensors. The return is averaged over 100 test episodes at the end of each training epoch, and shown upon division by 1000.}
%    \label{Convergence_Fig_2UAVs}
%\end{figure}

%\begin{figure}[t!]
    %\centering
%    \subfloat[Dataset of $5000$ rows]{\includegraphics[width=0.44\textwidth,trim={0.1 0 0 0},clip]{Figures/Conv_3_5000.eps} \label{Conv_3_5000}} 
%    \hskip -2.28ex
%    \subfloat[Dataset of $2000$ rows]{\includegraphics[width=0.44\textwidth,trim={0.1 0 0 0},clip]{Figures/Conv_3_2000.eps} \label{Conv_3_2000}} 
    %\hskip -2.28ex
%    \caption{Average test return as a function of the number of training epochs at the DT for an an environment of $3$ agents and $15$ sensors. The return is averaged over 100 test episodes at the end of each training epoch, and shown upon division by 1000.}
%    \label{Convergence_Fig_3UAVs}
%\end{figure}

\subsection{Numerical Results}
First, we show the simulation results of the proposed model via independent Q-learning compared to the baseline schemes. Then, we investigate the benefits of the joint training approach compared to independent training.

\subsubsection{Independent Learning}
Fig.~\ref{Convergence_Fig} shows the average test return, evaluated online using $100$ test episodes, for the policies obtained after a given number of training epochs. The figure thus reports the actual return obtained by the system as a function of the computational load, which increases with the number of training epochs.

We first observe that both MA-DIQN and MA-QR-DIQN, designed for online learning, fail to converge in the offline setting at hand. This well-known problem arises from overestimating Q-values corresponding to OOD actions in the offline dataset \cite{levine2020offline}. In contrast, conservative strategies designed for offline learning, namely MA-CIQL, MA-CIQR, and MA-CIQR-CVaR, exhibit an increasing average return as a function of the training epochs. In particular, the proposed MA-CIQR and MA-CIQR-CVaR provide the fastest convergence, needing around $30$ training epochs to reach the maximum return. In contrast, MA-CIQL shows slower convergence. This highlights the benefits of distributional RL in handling the inherent uncertainties arising in multi-agent systems from the environment and the actions of other agents~\cite{bdr2023}.

%\begin{figure}[t!]
%    \centering
%    \subfloat[$2$ agents]{\includegraphics[width=0.44\textwidth,trim={0.1 0 0 0},clip]{Figures/Ach_2_2000.eps} \label{Ach_2_2000}} 
%    \hskip -2.28ex
%    \subfloat[$3$ agents]{\includegraphics[width=0.44\textwidth,trim={0.1 0 0 0},clip]{Figures/Ach_3_2000.eps} \label{Ach_3_2000}} 
    %\hskip -2.28ex
%    \caption{Minimum sum-AoI as a function of the sum-power.}
%    \label{Achieveable_Reg}
%\end{figure}

\begin{figure}[t!]
    \centering    \includegraphics[width=1\columnwidth,trim={0 0 0 0},clip]{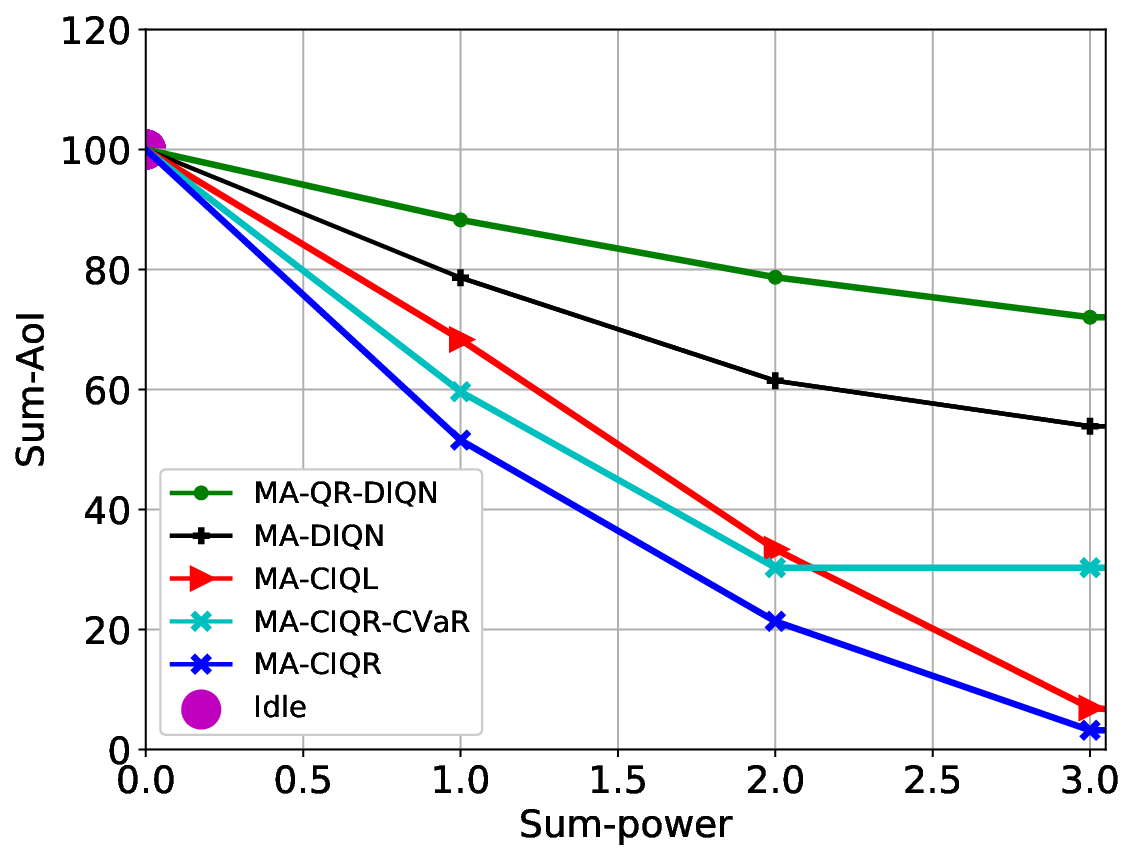} \vspace{2mm}
    \caption{Sum-AoI as a function of the sum-power using $\mathrm{P_{risk}}={\lambda}/{4}$ and $16 \%$ offline dataset for a system of $2$ UAVs serving $10$ sensors. %\MS{Better to have bigger Markers}
    % CQR achieves the least joint AoI and power among all the schemes.
    } 
    \vspace{0mm}
    \label{Achieveable_Reg}
\end{figure}

In Fig~\ref{Achieveable_Reg}, we report the optimal achievable trade-off between sum-AoI and sum-power consumption across the devices. This region is obtained by training the different schemes while sweeping the hyperparameter values $\lambda$. We recall that the hyperparameter $\lambda$ controls the weight of the power as compared to the AoI in the reward function (\ref{Reward_fn}). In particular, setting $\lambda \rightarrow 0$ minimizes the AoI only, resulting in a round-robin optimal policy. At the other extreme, setting a large value of $\lambda$ causes the UAV never to probe the devices, achieving the minimum power equal to zero, and the maximum AoI $A_{\max}=100$. This point is denoted as ``idle point'' the figure. The other curves represent the minimum sum-AoI achievable as a function of the sum-power.

From Fig.~\ref{Achieveable_Reg}, we observe that the proposed MA-CIQR always achieves the best age-power trade-off with the least age and sum-power consumption within all the curves. As in Fig. \ref{Convergence_Fig}, MA-DIQN and MA-QR-DIQN provide the worst performance due to their failure to handle the uncertainty arising from OOD actions. 

% \textcolor{blue}{It is noted that MA-CIQR-CVaR shows higher average sum-AoI (especially at higher average sum-power regions) compared to MA-CIQR due to the fact that MA-CIQR-CVaR scarifies the average performance slightly to avoid risky policies~\cite{ma2021conservative}.}

\begin{figure}[t!]
    \centering    \includegraphics[width=1\columnwidth,trim={0 0 0 0},clip]{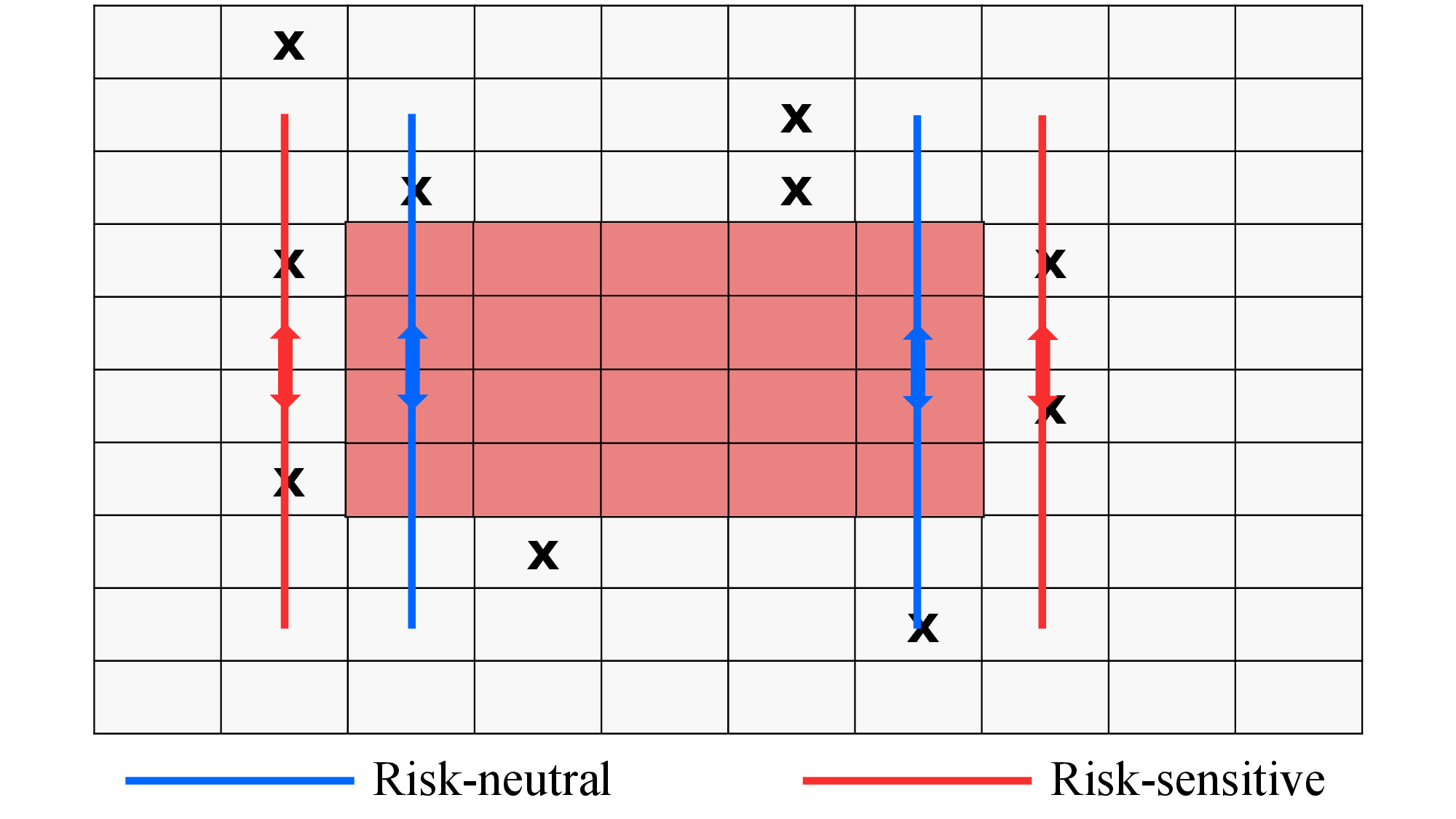} \vspace{2mm}
    \caption{A comparison between the trajectories of two UAVs using risk-neutral and risk-sensitive policies obtained via MA-CIQR and MA-CIQR-CVaR, respectively. Crosses represent the positions of the devices.
    % The risk-neutral agent may not take into account avoiding the risk sources in the environment, whereas the risk-sensitive agent avoids entering the risk region.
    } 
    \vspace{0mm}
    \label{Traj_Fig}
\end{figure}

In the next experiment, we investigate the capacity of the proposed risk-sensitive scheme MA-CIQR-CVaR to avoid risky trajectories. As a first illustration of this aspect, Fig. \ref{Traj_Fig} shows two examples of trajectories obtained via MA-CIQR and MA-CIQR-CVaR. It is observed that the risk-neutral policies obtained by MA-CIQR take shortcuts through the risk area, while the risk-sensitive trajectories obtained via MA-CIQR-CVaR avoid entering the risk area.

%\MS{is there a more interesting trajectory rather than a straight line ?}

\subsubsection{Centralized Learning}
\begin{figure}[t!]
    \centering
    \subfloat[$16 \%$ dataset size]{\includegraphics[width=0.48\textwidth,trim={0.1 0 0 0},clip]{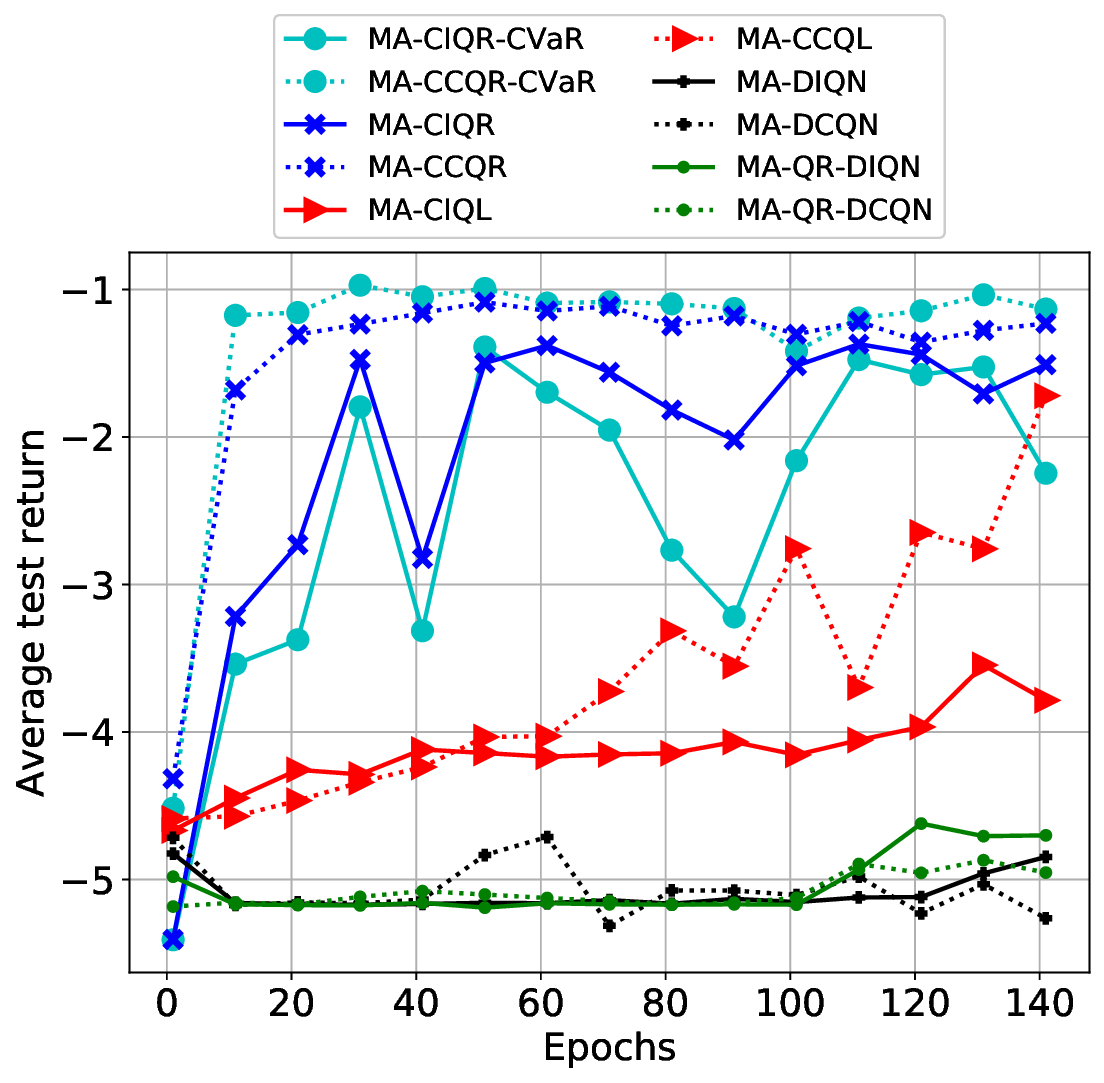} \label{Conv_2_5000}} 
    \hskip -2.28ex
    \subfloat[$6 \%$ dataset size]{\includegraphics[width=0.48\textwidth,trim={0.1 0 0 0},clip]{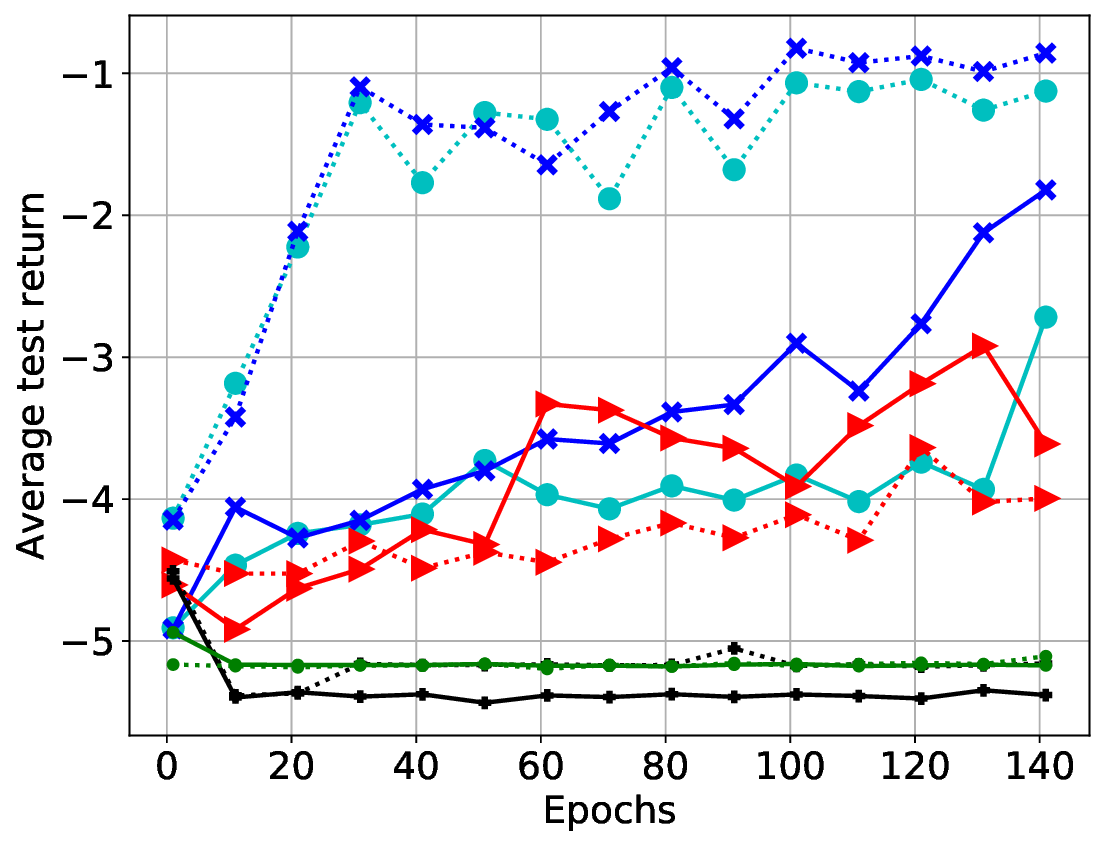} \label{Conv_2_2000}} 
    %\hskip -2.28ex
    \caption{Average test return as a function of the number of training epochs for penalty $\mathrm{P_{risk}}=300$ in a system of $3$ UAVs and $15$ sensors with data set size equal to (a) $16 \%$ and (b) $6 \%$. The return is averaged over $100$ test episodes at the end of each training epoch, and shown upon division by $1000$.}%\MS{I highly recommend reducing the number of points here. The figures are not well presented this way.}}
    \label{Convergence_Fig_2UAVs}
\end{figure}

Here, we compare the centralized training approach with independent learning. Fig.~\ref{Convergence_Fig_2UAVs} shows the average test return as a function of training epochs for an environment of $2$ agents and $10$ sensors. We use two offline datasets with different sizes, equal to $6 \%$ and $16 \%$ of the total transitions from the observed experience of online DQN agents. We increase the value of the risk region penalty to $\mathrm{P_{risk}}=300$ compared to the previous subsection experiments.

Fig.~\ref{Conv_2_5000} elucidates that the performance of the independent learning schemes is affected by increasing the risk penalty $\mathrm{P_{risk}}$. Specifically, MA-CIQL fails to reach convergence, while MA-CIQR and MA-CIQR-CVaR reach near-optimal performance but with a slower and less stable convergence than their centralized variants. However, as in Fig.~\ref{Conv_2_2000}, a significant performance gap is observed between the proposed independent schemes and their centralized counterpart for reduced dataset size. This joint training approach in coordinating between agents during training, requiring less data to obtain effective policies. Finally, we note the joint training approach did not enhance the performance of MA-DCQN and MA-QR-DCQN as these schemes are still heavily affected by the distributional shift in the offline setting. 

% We first observe that non-distributional schemes, i.e., MA-CIQL and MA-CCQL fail to reach convergence in the case of low dataset size, whereas only MA-CCQL reaches convergence in the case of using $16 \%$ data set size. This occurs due to the low number of data points ($16 \%$) used in the training. On the other hand, the proposed models, namely, MA-CIQR, MA-CIQR-CVaR, MA-CCQR, and MA-CCQR-CVaR have higher average test return as function of the training epochs. However, only the centralized training schemes, namely, MA-CCQR and MA-CCQR-CVaR reach convergence in the case of using $6 \%$ data set size. This reports that the CTDE schemes outperform the independent training as centralized training utilizes the information about the Q-function of all the agents prior to the decentralized execution.

%\begin{figure}[t!]
    %\centering    \includegraphics[width=1\columnwidth,trim={0 0 0 0},clip]{Figures/Conv_3_2000.eps} \vspace{2mm}
%    \caption{Average test return as a function of the number of training epochs at the DT for an an environment of $3$ agents and $15$ sensors. The return is averaged over 100 test episodes at the end of each training epoch, and shown upon division by 1000.
%    } 
%    \vspace{0mm}
%    \label{Convergence_Fig_2UAVs}
%\end{figure}

\begin{figure}[t!]
    \centering    \includegraphics[width=1\columnwidth,trim={0 0 0 0},clip]{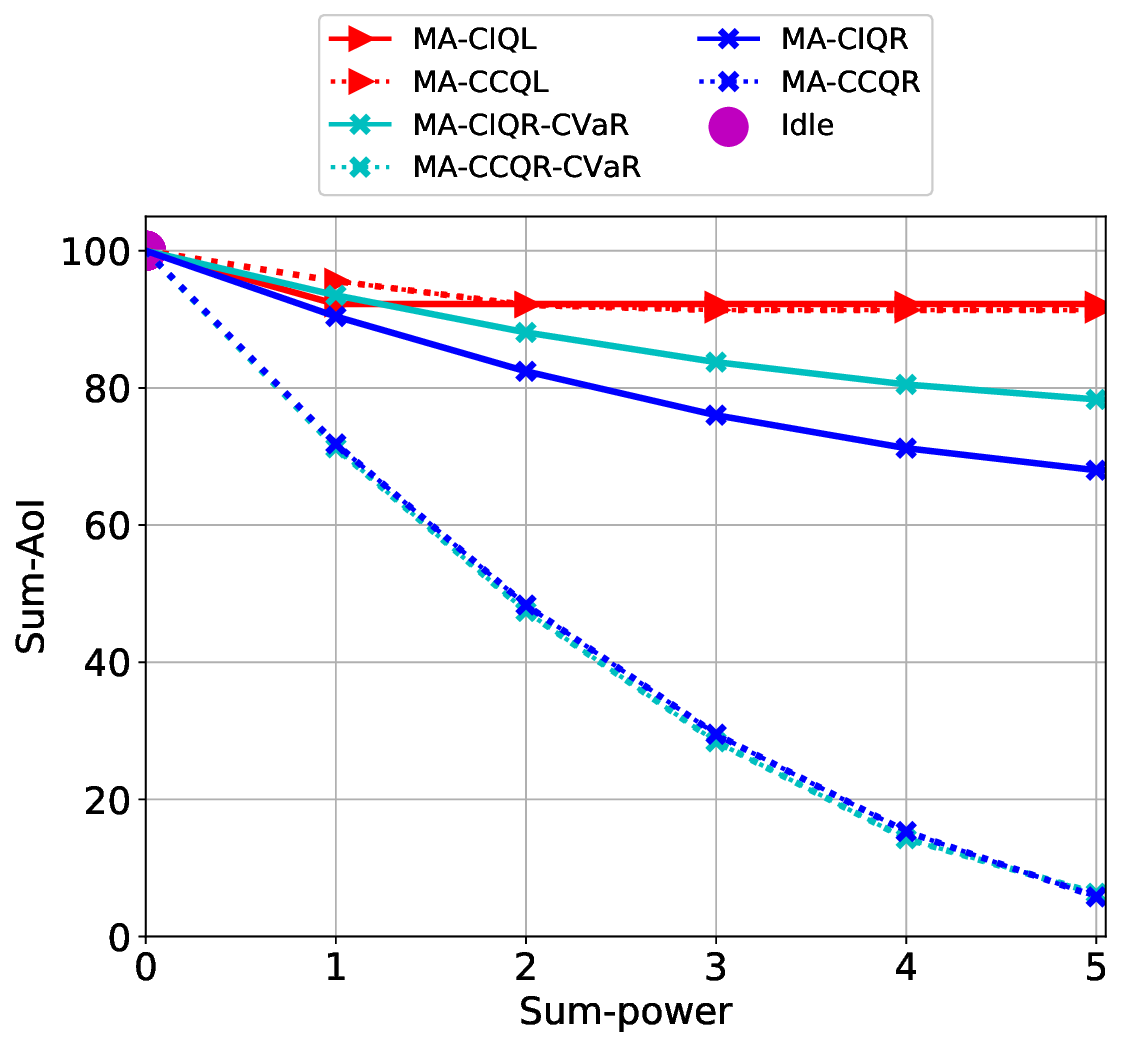} \vspace{2mm}
    \caption{Sum-AoI as a function of the sum-power for penalty $\mathrm{P_{risk}}={\lambda}/{2.5}$ in a system of $3$ UAVs and $15$ sensors (data set size equal to $6 \%$).} 
    \vspace{0mm}
    \label{Achieveable_Reg_2000}
\end{figure}

In a manner similar to Fig.~\ref{Achieveable_Reg}, Fig.~\ref{Achieveable_Reg_2000} shows the trade-off between sum-AoI and sum-power consumption for both independent and centralized training approaches for a system of $3$ agents serving $15$ sensors. Increasing the parameter $\lambda$ in~\eqref{Reward_fn} reduces the total power consumption at the expense of AoI. Here again, we observe a significant gain in performance for MA-CCQR and MA-CCQR-CVaR compared to their independent variants. In contrast, the non-distributional schemes, MA-CIQL and MA-CCQL, show similar results, as both perform poorly in this low data regime. We also observe that the average return performance of MA-CIQR-CVaR is worse than that of MA-CIQR, while MA-CCQR-CVaR provides a comparable performance as its risk-neutral counterpart MA-CCQR. This result suggests that, while producing low-risk trajectories, MA-CIQR-CVaR can yield lower average returns as compared to the risky trajectories of its risk-neutral counterpart, MA-CIQR. In contrast, thanks to the higher level of coordination between the agents in the joint training approach, MA-CCQR-CVaR can find low-risk trajectories while maintaining a comparable average return as MA-CCQR.

%\begin{figure}[t!]
%    \centering    \includegraphics[width=1\columnwidth,trim={0 0 0 0},clip]{Figures/Ach_3_2000.eps} \vspace{2mm}
%    \caption{Minimum sum-AoI as a function of the sum-power using $\mathrm{P_{risk}}={\lambda}/{4}$.
    % CQR achieves the least joint AoI and power among all the schemes.
%    } 
%    \vspace{0mm}
%    \label{Achieveable_Reg_2000}
%\end{figure}

Finally, to gain further insights into the comparison between MA-CCQR and MA-CCQR-CVaR, we leverage two metrics as in~\cite{ma2021conservative}, namely the \emph{percentage of violations} and the $\text{CVaR}_{0.15}$ return. The former is the percentage of timesteps at which one of the UAVs enters the risk region with respect to the total number of timesteps. In contrast, the $\text{CVaR}_{0.15}$ metric is the average return of the 15\% worst episodes. 

In Table~\ref{tab:COMP}, we report these two metrics, as well as the average return, with all returns normalized by 1000.
%{\color{blue} with all returns divided by 1000}.
Thanks to the ability of MA-CCQR-CVaR to learn how to avoid the risk region, this scheme has the lowest percentage of violations among all the schemes. In addition, it achieves the largest $\text{CVaR}_{0.15}$ return, with a small gain in terms of average return as compared to MA-CCQR. This demonstrates the advantages of the risk-sensitive design of policies.

%\MS{Would be good also to use this tool to evaluate the energy consumption and carbon footprint of each scheme:
%https://github.com/sb-ai-lab/Eco2AI
%Check Table 2 here as an example: https://ieeexplore.ieee.org/stamp/stamp.jsp?arnumber=10278721}
% \begin{table}[t!]
% \centering
% \caption{The quantitative evaluation of $100$ test episodes of two UAVs serving $10$ sensors with $\lambda = 500$ $\mathcal{P} = \frac{500}{4}$, $p_{risk} = 0.1$, and $\xi \: (\text{CVaR}) = 0.15$. CQR-CVaR is the only scheme who is risk-sensitive and the least scheme to enter the risk region.}
% \label{tab:COMP}
% \begin{tabular}{@{}c|ccc@{}}
% \toprule
% \textbf{Algorithm} & $\:$ Mean $\:$ & $\:$ CVaR $(0.15)$ $\:$ & $\:$ Violations $\:$ 

% \\ \midrule
% \vspace{0.01cm}\\

% DQN (online) & $-960.16$ & $-1232.34$ & $1193$ \\

% \vspace{0.01cm}\\
% \hline \\
% %\vspace{0.01cm}\\

% DQN (offline) & $-4143.89$ & $-4665.53$ & $2552$ \\

% \vspace{0.01cm}\\
% \hline \\
% %\vspace{0.01cm}\\

% QR-DQN  & $-5159.65$ & $-5402.09$ & $799$ \\

% \vspace{0.01cm}\\
% \hline \\
% %\vspace{0.01cm}\\

% CQL & $-1277.22$ & $-2872.14$ & $801$ \\

% \vspace{0.01cm}\\
% \hline \\
% %\vspace{0.01cm}\\

% CQR & $\boldsymbol{-782.07}$ & $-1017.56$ & $976$ \\

% \vspace{0.01cm}\\
% \hline \\
% %\vspace{0.01cm}\\

% CQR-CVaR & $-811.68$ & $\boldsymbol{-919.61}$ & $\boldsymbol{430}$ \\

% \vspace{0.01cm}\\

% \bottomrule
% \end{tabular}
% \end{table}

\begin{table}[t!]
\centering
 \caption{Performance evaluation over $100$ test episodes after $150$ training iterations for penalty $\mathrm{P_{risk}}={\lambda}/{2.5}$ in a system of $3$ UAVs and $15$ sensors (data set size equal to $6 \%$).
}
\label{tab:COMP}
\begin{tabular}{@{}cccc@{}}
\toprule
\textbf{Algorithm} & $\:$ Average return $\:$ & $\:$ $\text{CVaR}_{0.15}$ return $\:$ & $\:$ Violations $\:$ \\ \midrule
\midrule
MA-DQN (online) & $-1.5633$ & $-1.9611$ & $11.83\%$ \\
\hline 
MA-DCQN & $-4.8993$ & $-5.4930$ & $8.29\%$ \\
\hline 
MA-QR-DCQN  & $-4.2987$ & $-4.5743$ & $6.85\%$ \\
\hline 
MA-CCQL & $-3.8518$ & $-4.4695$ & $17.7\%$ \\
\hline 
MA-CCQR & $-1.4028$ & $-2.1775$ & $8.56\%$ \\
\hline 
MA-CCQR-CVaR & $\boldsymbol{-1.3641}$ & $\boldsymbol{-1.8513}$ & $\boldsymbol{5.83\%}$ \\
\bottomrule
\end{tabular}
\end{table}

% we evaluate each scheme across $100$ testing episodes and calculate the mean return, CVaR $\big(0.15\big)$ return (average over the worst $15$ episodes), and the number of violations (the number times an agent enters the risk region). In terms of the mean return, CQR outperforms the other schemes as it maximizes the average objective as illustrated in~\eqref{MARL_Objective_Average}.  CQR-CVaR learns to avoid the risk action; therefore, it has the least number of violations among all the schemes. In addition, on the extreme scenarios (averaging over the worst $15$ episodes), CQR-CVaR achieves the highest CVaR ($0.15$) return.

\section{Conclusions}\label{sec:conclusions} %\vspace{1mm}

In this paper, we developed a distributional and conservative offline MARL scheme for wireless systems. We considered optimizing the CVaR of the cumulative return to obtain risk-sensitive policies. We introduced two variants of the proposed scheme depending on the level of coordination between the agents during training. The proposed algorithms were applied to the trajectory optimization problem in UAV networks. Numerical results illustrate that the learned policies avoid risky trajectories more effectively and yield the best performance compared to the baseline MARL schemes. The proposed approach can be extended by considering online fine-tuning of the policies in the environment to handle the possible changes in the deployment environment compared to the one generating the offline dataset. Finally, the analysis of model-based offline MARL, which can leverage information about the physical system to learn a model of the environment, is left for future work.

%\section*{Acknowledgments} \vspace{1mm}
%This work is partially supported by Academy of Finland, 6G Flagship program (Grant no. 346208), and the European Commission through the Horizon Europe project Hexa-X (Grant Agreement no. 101015956). 

\bibliographystyle{IEEEtran}
\bibliography{IEEEabrv,references}

\begin{thebibliography}{10}
\providecommand{\url}[1]{#1}
\csname url@samestyle\endcsname
\providecommand{\newblock}{\relax}
\providecommand{\bibinfo}[2]{#2}
\providecommand{\BIBentrySTDinterwordspacing}{\spaceskip=0pt\relax}
\providecommand{\BIBentryALTinterwordstretchfactor}{4}
\providecommand{\BIBentryALTinterwordspacing}{\spaceskip=\fontdimen2\font plus
\BIBentryALTinterwordstretchfactor\fontdimen3\font minus \fontdimen4\font\relax}
\providecommand{\BIBforeignlanguage}[2]{{%
\expandafter\ifx\csname l@#1\endcsname\relax
\typeout{** WARNING: IEEEtran.bst: No hyphenation pattern has been}%
\typeout{** loaded for the language `#1'. Using the pattern for}%
\typeout{** the default language instead.}%
\else
\language=\csname l@#1\endcsname
\fi
#2}}
\providecommand{\BIBdecl}{\relax}
\BIBdecl

\bibitem{lavin2021simulation}
A.~Lavin, D.~Krakauer, H.~Zenil, J.~Gottschlich, T.~Mattson, J.~Brehmer, A.~Anandkumar, S.~Choudry, K.~Rocki, A.~G. Baydin \emph{et~al.}, ``Simulation intelligence: Towards a new generation of scientific methods,'' \emph{arXiv preprint arXiv:2112.03235}, 2021.

\bibitem{wang2020artificial}
C.-X. Wang, M.~Di~Renzo, S.~Stanczak, S.~Wang, and E.~G. Larsson, ``Artificial intelligence enabled wireless networking for 5{G} and beyond: Recent advances and future challenges,'' \emph{IEEE Wireless Communications}, vol.~27, no.~1, pp. 16--23, 2020.

\bibitem{chen2019artificial}
M.~Chen, U.~Challita, W.~Saad, C.~Yin, and M.~Debbah, ``Artificial neural networks-based machine learning for wireless networks: A tutorial,'' \emph{IEEE Communications Surveys \& Tutorials}, vol.~21, no.~4, pp. 3039--3071, 2019.

\bibitem{8714026}
N.~C. Luong, D.~T. Hoang, S.~Gong, D.~Niyato, P.~Wang, Y.-C. Liang, and D.~I. Kim, ``Applications of deep reinforcement learning in communications and networking: A survey,'' \emph{IEEE Communications Surveys \& Tutorials}, vol.~21, no.~4, pp. 3133--3174, 2019.

\bibitem{chen2021deep}
W.~Chen, X.~Qiu, T.~Cai, H.-N. Dai, Z.~Zheng, and Y.~Zhang, ``Deep reinforcement learning for internet of things: A comprehensive survey,'' \emph{IEEE Communications Surveys \& Tutorials}, vol.~23, no.~3, pp. 1659--1692, 2021.

\bibitem{marl-book}
\BIBentryALTinterwordspacing
S.~V. Albrecht, F.~Christianos, and L.~Sch\"afer, \emph{Multi-Agent Reinforcement Learning: Foundations and Modern Approaches}.\hskip 1em plus 0.5em minus 0.4em\relax MIT Press, 2024. [Online]. Available: \url{https://www.marl-book.com}
\BIBentrySTDinterwordspacing

\bibitem{levine2020offline}
S.~Levine, A.~Kumar, G.~Tucker, and J.~Fu, ``Offline reinforcement learning: Tutorial, review, and perspectives on open problems,'' \emph{arXiv preprint arXiv:2005.01643}, 2020.

\bibitem{bdr2023}
M.~G. Bellemare, W.~Dabney, and M.~Rowland, \emph{Distributional Reinforcement Learning}.\hskip 1em plus 0.5em minus 0.4em\relax MIT Press, 2023, \url{http://www.distributional-rl.org}.

\bibitem{kumar2020conservative}
\BIBentryALTinterwordspacing
A.~Kumar, A.~Zhou, G.~Tucker, and S.~Levine, ``Conservative {Q}-learning for offline reinforcement learning,'' in \emph{Advances in Neural Information Processing Systems}, vol.~33.\hskip 1em plus 0.5em minus 0.4em\relax Curran Associates, Inc., 2020, pp. 1179--1191. [Online]. Available: \url{https://proceedings.neurips.cc/paper_files/paper/2020/file/0d2b2061826a5df3221116a5085a6052-Paper.pdf}
\BIBentrySTDinterwordspacing

\bibitem{abd2019deep}
M.~A. Abd-Elmagid, A.~Ferdowsi, H.~S. Dhillon, and W.~Saad, ``Deep reinforcement learning for minimizing age-of-information in {UAV}-assisted networks,'' in \emph{2019 IEEE Global Communications Conference (GLOBECOM)}.\hskip 1em plus 0.5em minus 0.4em\relax IEEE, 2019, pp. 1--6.

\bibitem{samir2020age}
M.~Samir, C.~Assi, S.~Sharafeddine, D.~Ebrahimi, and A.~Ghrayeb, ``Age of information aware trajectory planning of {UAVs} in intelligent transportation systems: A deep learning approach,'' \emph{IEEE Transactions on Vehicular Technology}, vol.~69, no.~11, pp. 12\,382--12\,395, 2020.

\bibitem{eldeeb2022multi}
E.~Eldeeb, J.~M. de~Souza~Sant'Ana, D.~E. P{\'e}rez, M.~Shehab, N.~H. Mahmood, and H.~Alves, ``Multi-{UAV} path learning for age and power optimization in {IoT} with {UAV} battery recharge,'' \emph{IEEE Transactions on Vehicular Technology}, vol.~72, no.~4, pp. 5356--5360, 2022.

\bibitem{kidambi2020morel}
\BIBentryALTinterwordspacing
R.~Kidambi, A.~Rajeswaran, P.~Netrapalli, and T.~Joachims, ``Morel: Model-based offline reinforcement learning,'' in \emph{Advances in Neural Information Processing Systems}, vol.~33.\hskip 1em plus 0.5em minus 0.4em\relax Curran Associates, Inc., 2020, pp. 21\,810--21\,823. [Online]. Available: \url{https://proceedings.neurips.cc/paper_files/paper/2020/file/f7efa4f864ae9b88d43527f4b14f750f-Paper.pdf}
\BIBentrySTDinterwordspacing

\bibitem{ciosek2022imitation}
\BIBentryALTinterwordspacing
K.~Ciosek, ``Imitation learning by reinforcement learning,'' in \emph{International Conference on Learning Representations}, 2022. [Online]. Available: \url{https://openreview.net/forum?id=1zwleytEpYx}
\BIBentrySTDinterwordspacing

\bibitem{schulman2015trust}
J.~Schulman, S.~Levine, P.~Abbeel, M.~Jordan, and P.~Moritz, ``Trust region policy optimization,'' in \emph{International conference on machine learning}.\hskip 1em plus 0.5em minus 0.4em\relax PMLR, 2015, pp. 1889--1897.

\bibitem{yu2020mopo}
T.~Yu, G.~Thomas, L.~Yu, S.~Ermon, J.~Y. Zou, S.~Levine, C.~Finn, and T.~Ma, ``Mopo: Model-based offline policy optimization,'' \emph{Advances in Neural Information Processing Systems}, vol.~33, pp. 14\,129--14\,142, 2020.

\bibitem{Barde2024}
P.~Barde, J.~Foerster, D.~Nowrouzezahrai, and A.~Zhang, ``A model-based solution to the offline multi-agent reinforcement learning coordination problem,'' in \emph{the 23rd International Conference on Autonomous Agents and Multiagent Systems}.\hskip 1em plus 0.5em minus 0.4em\relax International Foundation for Autonomous Agents and Multiagent Systems, 2024.

\bibitem{yang2023offline}
K.~Yang, C.~Shi, C.~Shen, J.~Yang, S.-p. Yeh, and J.~J. Sydir, ``Offline reinforcement learning for wireless network optimization with mixture datasets,'' \emph{IEEE Transactions on Wireless Communications}, pp. 1--1, 2024.

\bibitem{bellemare2017distributional}
M.~G. Bellemare, W.~Dabney, and R.~Munos, ``A distributional perspective on reinforcement learning,'' in \emph{International conference on machine learning}.\hskip 1em plus 0.5em minus 0.4em\relax PMLR, 2017, pp. 449--458.

\bibitem{dabney2017distributional}
W.~Dabney, M.~Rowland, M.~Bellemare, and R.~Munos, ``Distributional reinforcement learning with quantile regression,'' in \emph{Proceedings of the AAAI Conference on Artificial Intelligence}, vol.~32, no.~1, 2018.

\bibitem{dabney2018implicit}
W.~Dabney, G.~Ostrovski, D.~Silver, and R.~Munos, ``Implicit quantile networks for distributional reinforcement learning,'' in \emph{International conference on machine learning}.\hskip 1em plus 0.5em minus 0.4em\relax PMLR, 2018, pp. 1096--1105.

\bibitem{rockafellar2000optimization}
R.~T. Rockafellar, S.~Uryasev \emph{et~al.}, ``Optimization of conditional value-at-risk,'' \emph{Journal of risk}, vol.~2, pp. 21--42, 2000.

\bibitem{lim2022distributional}
S.~H. Lim and I.~Malik, ``Distributional reinforcement learning for risk-sensitive policies,'' \emph{Advances in Neural Information Processing Systems}, vol.~35, pp. 30\,977--30\,989, 2022.

\bibitem{ma2021conservative}
\BIBentryALTinterwordspacing
Y.~Ma, D.~Jayaraman, and O.~Bastani, ``Conservative offline distributional reinforcement learning,'' in \emph{Advances in Neural Information Processing Systems}, vol.~34.\hskip 1em plus 0.5em minus 0.4em\relax Curran Associates, Inc., 2021, pp. 19\,235--19\,247. [Online]. Available: \url{https://proceedings.neurips.cc/paper_files/paper/2021/file/a05d886123a54de3ca4b0985b718fb9b-Paper.pdf}
\BIBentrySTDinterwordspacing

\bibitem{jiang2023offline}
J.~Jiang and Z.~Lu, ``Offline decentralized multi-agent reinforcement learning.'' in \emph{ECAI}, 2023, pp. 1148--1155.

\bibitem{pan2022plan}
L.~Pan, L.~Huang, T.~Ma, and H.~Xu, ``Plan better amid conservatism: Offline multi-agent reinforcement learning with actor rectification,'' in \emph{International Conference on Machine Learning}.\hskip 1em plus 0.5em minus 0.4em\relax PMLR, 2022, pp. 17\,221--17\,237.

\bibitem{shao2023counterfactual}
\BIBentryALTinterwordspacing
J.~Shao, Y.~Qu, C.~Chen, H.~Zhang, and X.~Ji, ``Counterfactual conservative q learning for offline multi-agent reinforcement learning,'' in \emph{Advances in Neural Information Processing Systems}, vol.~36.\hskip 1em plus 0.5em minus 0.4em\relax Curran Associates, Inc., 2023, pp. 77\,290--77\,312. [Online]. Available: \url{https://proceedings.neurips.cc/paper_files/paper/2023/file/f3f2ff9579ba6deeb89caa2fe1f0b99c-Paper-Conference.pdf}
\BIBentrySTDinterwordspacing

\bibitem{wang2023offline}
\BIBentryALTinterwordspacing
X.~Wang, H.~Xu, Y.~Zheng, and X.~Zhan, ``Offline multi-agent reinforcement learning with implicit global-to-local value regularization,'' in \emph{Advances in Neural Information Processing Systems}, vol.~36.\hskip 1em plus 0.5em minus 0.4em\relax Curran Associates, Inc., 2023, pp. 52\,413--52\,429. [Online]. Available: \url{https://proceedings.neurips.cc/paper_files/paper/2023/file/a46c84276e3a4249ab7dbf3e069baf7f-Paper-Conference.pdf}
\BIBentrySTDinterwordspacing

\bibitem{yang2021believe}
\BIBentryALTinterwordspacing
Y.~Yang, X.~Ma, C.~Li, Z.~Zheng, Q.~Zhang, G.~Huang, J.~Yang, and Q.~Zhao, ``Believe what you see: Implicit constraint approach for offline multi-agent reinforcement learning,'' in \emph{Advances in Neural Information Processing Systems}, vol.~34.\hskip 1em plus 0.5em minus 0.4em\relax Curran Associates, Inc., 2021, pp. 10\,299--10\,312. [Online]. Available: \url{https://proceedings.neurips.cc/paper_files/paper/2021/file/550a141f12de6341fba65b0ad0433500-Paper.pdf}
\BIBentrySTDinterwordspacing

\bibitem{eldeeb2023traffic}
E.~Eldeeb, M.~Shehab, and H.~Alves, ``Traffic learning and proactive {UAV} trajectory planning for data uplink in markovian {IoT} models,'' \emph{IEEE Internet of Things Journal}, vol.~11, no.~8, pp. 13\,496--13\,508, 2024.

\bibitem{9322539}
F.~Wu, H.~Zhang, J.~Wu, L.~Song, Z.~Han, and H.~V. Poor, ``{AoI} minimization for {UAV}-to-device underlay communication by multi-agent deep reinforcement learning,'' in \emph{GLOBECOM 2020 - 2020 IEEE Global Communications Conference}, 2020, pp. 1--6.

\bibitem{8807386}
J.~Cui, Y.~Liu, and A.~Nallanathan, ``Multi-agent reinforcement learning-based resource allocation for {UAV} networks,'' \emph{IEEE Transactions on Wireless Communications}, vol.~19, no.~2, pp. 729--743, 2020.

\bibitem{8792117}
Y.~S. Nasir and D.~Guo, ``Multi-agent deep reinforcement learning for dynamic power allocation in wireless networks,'' \emph{IEEE Journal on Selected Areas in Communications}, vol.~37, no.~10, pp. 2239--2250, 2019.

\bibitem{naderializadeh2021resource}
N.~Naderializadeh, J.~J. Sydir, M.~Simsek, and H.~Nikopour, ``Resource management in wireless networks via multi-agent deep reinforcement learning,'' \emph{IEEE Transactions on Wireless Communications}, vol.~20, no.~6, pp. 3507--3523, 2021.

\bibitem{xu2023digital}
C.~Xu, Z.~Tang, H.~Yu, P.~Zeng, and L.~Kong, ``Digital twin-driven collaborative scheduling for heterogeneous task and edge-end resource via multi-agent deep reinforcement learning,'' \emph{IEEE Journal on Selected Areas in Communications}, vol.~41, no.~10, pp. 3056--3069, 2023.

\bibitem{zhang2021millimeter}
Q.~Zhang, W.~Saad, and M.~Bennis, ``Millimeter wave communications with an intelligent reflector: Performance optimization and distributional reinforcement learning,'' \emph{IEEE Transactions on Wireless Communications}, vol.~21, no.~3, pp. 1836--1850, 2021.

\bibitem{zhang2020distributional}
------, ``Distributional reinforcement learning for mmwave communications with intelligent reflectors on a uav,'' in \emph{GLOBECOM 2020-2020 IEEE Global Communications Conference}.\hskip 1em plus 0.5em minus 0.4em\relax IEEE, 2020, pp. 1--6.

\bibitem{hua2019gan}
Y.~Hua, R.~Li, Z.~Zhao, X.~Chen, and H.~Zhang, ``Gan-powered deep distributional reinforcement learning for resource management in network slicing,'' \emph{IEEE Journal on Selected Areas in Communications}, vol.~38, no.~2, pp. 334--349, 2019.

\bibitem{sunehag2017value}
P.~Sunehag, G.~Lever, A.~Gruslys, W.~M. Czarnecki, V.~Zambaldi, M.~Jaderberg, M.~Lanctot, N.~Sonnerat, J.~Z. Leibo, K.~Tuyls \emph{et~al.}, ``Value-decomposition networks for cooperative multi-agent learning,'' \emph{arXiv preprint arXiv:1706.05296}, 2017.

\bibitem{lyu2024centralizedcriticsmultiagentreinforcement}
\BIBentryALTinterwordspacing
X.~Lyu, A.~Baisero, Y.~Xiao, B.~Daley, and C.~Amato, ``On centralized critics in multi-agent reinforcement learning,'' 2024. [Online]. Available: \url{https://arxiv.org/abs/2408.14597}
\BIBentrySTDinterwordspacing

\bibitem{lowe2020multiagent}
\BIBentryALTinterwordspacing
R.~Lowe, Y.~WU, A.~Tamar, J.~Harb, O.~Pieter~Abbeel, and I.~Mordatch, ``Multi-agent actor-critic for mixed cooperative-competitive environments,'' in \emph{Advances in Neural Information Processing Systems}, vol.~30.\hskip 1em plus 0.5em minus 0.4em\relax Curran Associates, Inc., 2017. [Online]. Available: \url{https://proceedings.neurips.cc/paper_files/paper/2017/file/68a9750337a418a86fe06c1991a1d64c-Paper.pdf}
\BIBentrySTDinterwordspacing

\bibitem{bellman1966dynamic}
R.~Bellman, ``Dynamic programming,'' \emph{Science}, vol. 153, no. 3731, pp. 34--37, 1966.

\bibitem{xu2023offline}
H.~Xu, L.~Jiang, J.~Li, Z.~Yang, Z.~Wang, V.~W.~K. Chan, and X.~Zhan, ``Offline rl with no ood actions: In-sample learning via implicit value regularization,'' \emph{arXiv preprint arXiv:2303.15810}, 2023.

\bibitem{9815722}
E.~Eldeeb, D.~E. P\'erez, J.~Michel~de Souza~Sant'Ana, M.~Shehab, N.~H. Mahmood, H.~Alves, and M.~Latva-Aho, ``A learning-based trajectory planning of multiple {UAVs} for {AoI} minimization in {IoT} networks,'' in \emph{2022 Joint European Conference on Networks and Communications \& 6G Summit (EuCNC/6G Summit)}, 2022, pp. 172--177.

\bibitem{9845353}
E.~Eldeeb, M.~Shehab, A.~E. Kal\o~r, P.~Popovski, and H.~Alves, ``Traffic prediction and fast uplink for hidden markov {IoT} models,'' \emph{IEEE Internet of Things Journal}, vol.~9, no.~18, pp. 17\,172--17\,184, 2022.

\bibitem{AoI}
A.~Kosta, N.~Pappas, and V.~Angelakis, ``Age of information: A new concept, metric, and tool,'' \emph{Foundations and Trends in Networking, Now Publishers, Inc.}, 2017.

\bibitem{10293964}
E.~Eldeeb, M.~Shehab, and H.~Alves, ``Age minimization in massive {IoT} via {UAV} swarm: A multi-agent reinforcement learning approach,'' in \emph{2023 IEEE 34th Annual International Symposium on Personal, Indoor and Mobile Radio Communications (PIMRC)}, 2023, pp. 1--6.

\bibitem{tampuu2017multiagent}
A.~Tampuu, T.~Matiisen, D.~Kodelja, I.~Kuzovkin, K.~Korjus, J.~Aru, J.~Aru, and R.~Vicente, ``Multiagent cooperation and competition with deep reinforcement learning,'' \emph{PloS one}, vol.~12, no.~4, p. e0172395, 2017.

\end{thebibliography}
\end{document}